\documentclass[10ptb]{article} 
\usepackage[margin=1.5in]{geometry}
\usepackage{hyperref}
\usepackage{url}
\usepackage{amsmath}
\usepackage{amsfonts}
\usepackage{amssymb}

\usepackage[pdftex]{graphicx}

\usepackage{color}

\usepackage{xfrac}

\usepackage{fancyhdr}
\title{Quantum Clustering and  Gaussian Mixtures}
\author{Mahajabin Rahman \and Davi Geiger}
\date{}

\begin{document}

\maketitle
\thispagestyle{empty}

\begin{abstract}

The mixture of Gaussian distributions, a soft version of {\em k}-means ( \cite{Bilmes1998}), is considered a state-of-the-art clustering algorithm. It is widely used in computer vision for selecting classes, e.g., color\cite{CELENK1990, Malik1998, Chen2008},texture\cite{Malik1998, kim2007}, shapes \cite{Srivastava2005,Likas2003}. In this algorithm, each class is  described by a Gaussian distribution, defined by its mean and covariance. The data is described by a weighted sum of these Gaussian distributions.

We propose a new method, inspired by quantum interference in physics \cite{Feynman1971}. Instead of modeling each class distribution directly, we model a {\bf class wave function} such that its magnitude square is the class Gaussian  distribution.  We then mix the {\bf class wave functions} to create the {\bf mixture wave function}. The final mixture distribution is then the magnitude square of the {\bf mixture wave function}.  As a result, we observe the quantum class interference phenomena, not present in the Gaussian mixture model. We show that the quantum method outperforms the Gaussian mixture method in every aspect of the estimations. It provides more accurate estimations of all distribution parameters, with much less fluctuations,  and  it is  also more robust to data deformations from the Gaussian assumptions. We illustrate our method for color segmentation as an example application.
\end{abstract}

\section{Introduction}
\label{sec:introduction}

\vspace{0.5cm}

Given a large set of non labeled data,  being able to separate it in clusters is of great interest in computer vision (e.g.,  \cite{Bilmes1998,CELENK1990, Malik1998, Chen2008, kim2007, Srivastava2005,Likas2003}.) 
Clustering may also be interpreted in a Bayesian method as the data likelihood model.  When combined with the prior model on the data (such as encouraging neighbors  to be labeled similarly), one forms the posterior probability. 

This paper is solely focused on the clustering model and in particular we focus on data generated from multiple Guassian distributions, each one characterized by mean and covariances. The Gaussian Mixture Model (GMM)  is a standard  model to describe such data and combined with  the expectation-maximization algorithm  is the standard approach to recover  the Gaussian parameters as well as the number of points per class (see \cite{Bilmes1998}).    Inspired by quantum methods, e.g.,  \cite{Feynman1971},  we reformulate this classical model (GMM) and produce  a quantum model.  Using  the theory of expectation-maximization we recover the Gaussian parameters as well as the number of points per class.   

The inspiration for our proposed method comes from the wave interference phenomena and previous work \cite{ICIP2015}. In the classical model, as one mixes distributions (such as a Gaussian mixture - GMM), no probabilities are canceled. By formulating the distributions in terms of wave functions, such that their magnitude square describe the probability distribution, we allow their mixtures to interfere. The effect is the cancellation of probabilities, which allows us to identify the clusters of data more accurately.

\section{Classical Clustering}
\label{sec:classical}
The input data is given by a set of N points $\{ p_1, p_2, ...,p_N\}$. 
If we are in two dimensions, our default point is, unless specified,  $p_i = (x_i, y_i)$ and if we are in three dimensions $p_i^{3D} = (x_i, y_i, z_i)$.

Given $K$ classes, each class data is produced by a Gaussian distribution with center $\mu_k$ and Covariance ${\bf C}_k$, with $k=1,...,K$, i.e., thus the probability of $p_i$ given class k is the following: 

\[P(p_{i}|k, \mu_k, {\bf C}_k)=\frac{1}{Z_k} \,  \,  e^{- \, \frac{1}{2} \, (p_i -\mu_k)\, {\bf C}_k^{-1} \, (p_i-\mu_k) }\, .\]
where $Z_k= (2\, \pi)^{d/2} \, |{\bf C}_k|^{1/2}$ is a normalization factor  with $ |{\bf C}_k|^{1/2}= \sqrt{det\, {\bf C}_k}$ and $d$ is the data dimensionality.

Suppose from the set of $N$ points originated from $K$ classes,  there are $N_k$ points per class ($N=\sum_{k=1}^{K} N_k$). Since $N_k$ is assumed to be  unknown, according to the Gaussian mixture method,  the final distribution is given by a weighted sum of each class distribution of all $N$ points

\begin{eqnarray}
P(p_i|\{\theta_k\}) &= & \sum_{k=1}^{K} P(p_i, k|\{\theta_k\})= \sum_{k=1}^{K} P(k|\{\theta_k\}) \, P(p_i|k, \{\theta_k\})
\nonumber \\
&=&\sum_{k=1}^{K}\left[   P(k|\{\theta_k\})  \, \frac{1}{\sqrt{(2\, \pi)^{d} \, |{\bf C}_k|}}\, e^{-  \, (p_i -\mu_k)\, {\bf C}_k^{-1} \, (p_i-\mu_k) } \right ]
\label{eq:classical-distribution}
\end{eqnarray}

\noindent where  $P(k|\{\theta_k\})$ is the prior probability that a point belongs to class $k$, and $\theta_k=\{{\bf C}_k, \mu_k\} $ are the Gaussian parameters of class $k$. The clustering problem is to find $P(p_i|\{\theta_k\})$ from the data, i.e., to extract the parameters $\theta=\{ P(k|\{\theta_k\}), \theta_k ; k=1, ...,K\}=\{ P(k|\{\theta_k\}), {\bf C}_k, \mu_k ; k=1, ...,K\}$ from the data. A standard method to extract these parameters is based on the EM method. 

\subsection{EM Method}
In appendix \ref{sec:Derivation-EM}, we show a brief derivation of the EM method that leads to 
 the iteration procedure $t=1, ..., T$ with the following two steps

\paragraph{E-step:}

\begin{eqnarray}
Q_i^{t}(k) &=&P(k| p_i  ,\{ \theta_k^{t-1} \}) 
=\frac{P(p_i,k|\{ \theta_k^{t-1} \})}{ P(p_i|\{ \theta_k^{t-1} \})}
=\frac{P(p_i,k|\{ \theta_k^{t-1} \})}{\sum_l P(p_i,l|\{ \theta_k^{t-1} \})}
\label{eq:E-step}
\end{eqnarray}

For the Gaussian mixture, we get
\begin{eqnarray}
Q_i^t(k)\equiv Q_{i,k}^t &=&  \frac{P^{t-1}(k|\{ \theta_k^{t-1} \})\, \frac{1}{Z_k^{t-1}} \,  e^{-  \, (p_i -\mu_k^{t-1})\, ({\bf C}_k^{-1})^{t-1} \, (p_i-\mu_k^{t-1}) }}{\sum_{l=1}^{K} P^{t-1}(l|\{ \theta_k^{t-1} \})\, \frac{1}{Z_l^{t-1}} \,  e^{-  \, (p_i -\mu_l^{t-1})\, ({\bf C}_l^{-1})^{t-1} \, (p_i-\mu_l^{t-1}) } } 
\label{eq:E-step-Gaussian-classical}
\end{eqnarray}

Followed by

\paragraph{M-Step:} The maximization of 
  \begin{eqnarray} 
[P^{t}(k), \theta_k^{t}]& = & arg \max_{P(k), \{\theta_k\}} \sum_i  \sum_k Q_i^{t}(k) \, \log \frac{ P( p_i, k|\{ \theta_k^{t-1} \}}{Q_i^{t}(k)} \nonumber \\
&=& arg \max_{P(k), \{\theta_k\}} \sum_i  \sum_k Q_i^{t}(k) \, \log  P( p_i, k|\{ \theta_k^{t-1} \})
\label{eq:optimization-criteria}
\end{eqnarray}

For the Gaussian mixture ($\theta_k=({\bf C}_k, \mu_k)$) we get

\begin{eqnarray}
N_k^t &=& \sum_i Q_{i,k}^t \nonumber \\
&\Downarrow&  N_k^t - {\rm estimated \, number \, of \, points\, , \, \, at \, step} \, t,  \, {\rm associated\,  to \, class } \, k\nonumber \\
P^{t}(k|\{ \theta_k^{t-1} \}) &=& <P^t(k|p_i,\{ \theta_k^{t-1}\})>_N =\frac{N_k^t}{\sum_{l=1}^K N_k^t}
\quad \nonumber \\
\mu_k^t &=&\frac{1}{N_k^t}\sum_{i=1}^{N} Q_{i,k}^t \, p_i \quad \nonumber \\
{\bf C}_k^{t}&=&\frac{1}{N_k^t}\sum_{i=1}^{N} Q_{i,k}^t \, (p_i - \mu_k^t) (p_i - \mu_k^t)^T
\label{eq:M-step}
\end{eqnarray}
where $< f(p_i) >_N$ indicates the expectation of the function $f$ over all $N$ points $\{ p_i \}$ (with respect to $P( p_i, k|\{ \theta_k^{t-1} \})$) . The Gaussian normalization $Z_k= \sqrt{(2\, \pi)^{d} \, |{\bf C}_k|}$ is also required to be computed.

\section{Quantum Clustering}
We introduce a wave function to describe the system of points. It is such that the probability associated to the system of points is the magnitude square of the wave function. The wave function associated to a point $p_i$  from a class $k$ is given by 


\[ \psi_k(p_i|\theta_k)\equiv \frac{1}{\sqrt{Z_k}}\, e^{- \frac{1}{4} \, (p_i-\mu_k)\, {\bf C}_k^{-1} \, (p_i-\mu_k) }\, e^{-i\, \phi_k }\, , \]
where $Z_k= (2\, \pi)^{d/2} \, |{\bf C}_k|^{1/2}$. From this definition we then recover the classical probability distribution for a given class 
$P(p_i|k,\theta_k)=|\psi_k(p_i|\theta_k)|^2= \frac{1}{Z_k} \, e^{-  \, \frac{1}{2} \, (p_i-\mu_k)\, {\bf C}_k \, (p_i-\mu_k) }$. The new phase factor, $ e^{-i\, \phi_k }$,  does not alter the class distribution, but will have its impact in the mixture case.

We  extend the quantum method when the data is described by a mixture of classes.  Analogously to the mixture of Gaussians, we now mix quantum waves to produce the final wave, i.e., 

\begin{eqnarray} 
\psi(p_i|\{\alpha_k,\theta_k\}) =\sum_{k=1}^K \alpha_k \, \psi_k(p_i|\theta_k) 
=  \sum_{k=1}^K \left[  \frac{\alpha_k}{\sqrt{Z_k}} \,\, e^{- \frac{1}{4} \, (p_i -\mu_k)\, {\bf C}_k^{-1}  \, (p_i-\mu_k) }\, e^{-i\, \phi_k  }
\right ] 
\label{eq:mixture-waves}
\end{eqnarray}
The final  probability $P(p_i)$ (short for $P(p_i|\{\alpha_k,\theta_k\})$, with $\theta_k=(\mu_k,\, {\bf C}_k, \phi_k)$,  is given by

\begin{eqnarray} 
P(p_i) &=&  |\psi(p_i|\{\alpha_k,\theta_k\})|^2 \qquad \qquad \qquad \qquad \qquad 
\\
&= &  \sum_{k=1}^K  \, \Bigg [
 \frac{\alpha_k}{\sqrt{Z_k}} \, \,  e^{-  \frac{1}{4}\, (p_i -\mu_k)\, {\bf C}_k \, (p_i-\mu_k) } \,
\left.  \sum_{l=1}^K  \, \frac{\alpha_l^*}{\sqrt{Z_l}} \, \,  \cos \phi_{l,k}(p_i) \, e^{-  \frac{1}{4}\, (p_i -\mu_l)\, {\bf C}_l \, (p_i-\mu_l) }  \, \right ]
\nonumber 
\label{eq:final-quantum}
\end{eqnarray}
where $\alpha_l^*$ is the conjugate of $\alpha_l$ ,  and 
\begin{equation}
\phi_{l,k}= \phi_k -\phi_l 
\label{eq:phase-quantum}
\end{equation}  
and  we must require that the set $\{\alpha_k\}$ satisfy $1=\sum_i P(p_i)$, 
The final probability $P(p_i|\{\alpha_k,\theta_k\})$ exhibits wave interference phenomena, not present in the classical clustering formulation,  due to  $  \cos \phi_{l,k}$ for $l\ne k$.

Following the Bayesian interpretation, $P(p_i|\{\alpha_k,\theta_k\}) = \sum_k P(p_i,k|\{\alpha_k,\theta_k\})$, where $\theta_k=(\mu_k, {\bf C}_k,  \phi_k)$, the join distribution $P(p_i,k|\{\alpha_k,\theta_k\})$ is derived from equation \ref{eq:final-quantum}

\begin{eqnarray} 
P(p_i,k|\{\alpha_k,\theta_k\}) 
&= &  \frac{\alpha_k}{\sqrt{Z_k}}\, \,  e^{-  \frac{1}{4}\, (p_i -\mu_k)\, {\bf C}_k \, (p_i-\mu_k) } \,
  \sum_{l=1}^K  \, \frac{\alpha_l^*}{\sqrt{Z_l}} \, \,  \cos \phi_{l,k}(p_i) \, e^{-  \frac{1}{4}\, (p_i -\mu_l)\, {\bf C}_l \, (p_i-\mu_l) } 
 \nonumber \\
 && {\rm Defining \, the\,  unnormalized \, Gaussian \,  model:\, }
 \nonumber \\
 && G_{i,k}=\frac{1}{\sqrt{Z_k}}\,  e^{-  \frac{1}{4}\, (p_i -\mu_k)\, {\bf C}_k^{-1} \, (p_i-\mu_k) }\quad ({\rm where} \, G_{i,k}^2 \, {\rm is \, normalized})\nonumber \\
 &= &
 \alpha_k \,  G_{i,k} \,
 \left (\sum_{l=1}^K  \, \alpha_l^* \,  \cos \phi_{l,k} \,  G_{i,l}\right )
\label{eq:joint-distribution-quantum}
\end{eqnarray}
 with the constraint on $\{\alpha_k\}$ that $1=\sum_i \sum_k P(p_i,k|\{\alpha_k,\theta_k\}) $.
 
\paragraph{Classical Case:} For $   \phi_{l,k}=\frac{\pi}{2}\,\, \forall \, l\ne k$, we recover the classical case. This can only happen if we have just two classes, otherwise for at least one pair of classes the phase difference will not be $\frac{\pi}{2}$.

\subsection{EM Method for Quantum Clustering}

\paragraph{E-step:} From equation \ref{eq:E-step}

\begin{eqnarray}
Q_i^{t}(k) &=&P(k| p_i  ,\{ \theta_k^{t-1} \}) 
=\frac{P(p_i,k|\{ \theta_k^{t-1} \})}{ P(p_i|\{ \theta_k^{t-1} \})}
=\frac{P(p_i,k|\{ \theta_k^{t-1} \})}{\sum_m P(p_i,m|\{ \theta_m^{t-1} \})}
\nonumber \\
&=&\left [ \frac{ \alpha_k \,  G_{i,k} \,
 \left (\sum_{l=1}^K  \, \alpha_l^* \,  \cos \phi_{l,k} \,  G_{i,l}\right )}{ \sum_{m}\alpha_m \,  G_{i,m} \,
 \left (\sum_{l=1}^K  \, \alpha_l^* \,  \cos \phi_{l,m} \,  G_{i,l}\right )}\right ]^{t-1}
\label{eq:E-quantum-step}
\end{eqnarray}

Note that, like in the classical case, the estimated number  of  points, \,  at  step  $t$,   associated\,  to \, class $ \, k$ is given by
\begin{eqnarray}
N_k^t &=& \sum_{i=1}^N P(k| p_i  ,\{ \theta_k^{t-1} \}) =\sum_{i=1}^N Q_{i,k}^t 
\label{eq:number-points-quantum}
\end{eqnarray}

\paragraph{M-Step:} The maximization of 
  \begin{eqnarray} 
(\alpha_k^{t},\theta_k^{t})& = & \arg \max_{\{\alpha_k,\theta_k\}} \sum_i  \sum_k Q_i^{t}(k) \, \log \frac{ P( p_i, k|\{ \theta_k^{t-1} \})}{Q_i^{t}(k)} \nonumber \\
&=& \arg \max_{\{\alpha_k,\theta_k\}} \sum_i  \sum_k Q_i^{t}(k) \, \log  P( p_i, k|\{ \theta_k^{t-1} \})
\label{eq:optimization-criteria}
\end{eqnarray}
where $\theta_k=(\mu_k, {\bf C}_k,  \phi_k)$.  
  
\subsection{Two Classes and Constraints}

For a two class model, $k=1,2$, without loss of generality, we can  consider  $\alpha_1, \alpha_2 \in \mathbb{R}$ since their possible phase is already absorbed by  $\phi_1$ and $\phi_2$. Then,  

\begin{eqnarray}
 \begin{matrix}
P(p_i,1|\{\alpha_k, \theta_k\}) &=&
\alpha_{1}^2 \,  \, G_{i,1}^2 
 + \alpha_{1} \, \alpha_{2} \, G_{i,1} \, G_{i,2} \, \cos \phi
 \\
 P(p_i,2|\{\alpha_k, \theta_k\}) &=&
\alpha_{2}^2 \,  \, G_{i,2}^2 
 + \alpha_{1} \, \alpha_{2} \, G_{i,1} \, G_{i,2} \, \cos \phi \\ \\
 &{\rm with } & \cos \phi = \cos \phi_{1,2} = \cos \phi_{2,1}\\ \\
\end{matrix}
\label{eq:pik-two-classes}
\end{eqnarray}

\paragraph {Constraints:} The probability $P( p_i, k|\{ \theta_k^{t-1} \})$ must satisfy 
 $$0\le P( p_i, k|\{ \theta_k^{t-1} \})\le 1$$
 and
 $$1=\sum_i \sum_k P( p_i, k|\{ \theta_k^{t-1} \}) \,\,  \Rightarrow \,  \, 1=\alpha_1^2 + \alpha_2^2+  2\, \alpha_1 \, \alpha_2\, \cos \phi \, \sum_i G_{i,1} \, G_{i,2}   \, .$$

\paragraph{E-step:} Replacing  $\cos \phi=\frac{1-\alpha_1^2 - \alpha_2^2}{2 \, \alpha_1 \, \alpha_2\,  \sum_i G_{i,1} \, G_{i,2}  }$  into  equation \ref{eq:E-quantum-step} and the E-step becomes

\begin{eqnarray}
Q_i^{t}(1) 
&=&  \left [ \frac{\alpha_1^2 \,  G_{i,1}^2 + \frac{1}{2}\, (\alpha\, o)_i }{ \alpha_{1}^2 \,  G_{i,1}^2 
 +  \alpha_{2}^2 \,  G_{i,2}^2 
 + \, (\alpha\, o)_i } \right ]^{t-1}
 \nonumber \\
 Q_i^{t}(2) 
&=&  \left [ \frac{\alpha_2^2 \,  G_{i,2}^2 + \, \frac{1}{2}\, (\alpha\, o)_i }{ \alpha_{1}^2 \,  G_{i,1}^2 
 +  \alpha_{2}^2 \,  G_{i,2}^2 
 + \, (\alpha\, o)_i  }
 \right ]^{t-1} \quad 
\label{eq:E-quantum-step-two-nocos}
\end{eqnarray}
where 
\begin{eqnarray}
o_i=\frac{G_{i,1}\, G_{i,2} }{ \sum_i G_{i,1} \, G_{i,2}} \quad {\rm and} \quad (\alpha\, o)_i = (1-\alpha_1^2 - \alpha_2^2)\, o_i
\label{eq:overlap-measures}
\end{eqnarray}
gives measures of overlap of the distributions. 
\paragraph{M-step:} Replacing  $\cos \phi=\frac{1-\alpha_1^2 - \alpha_2^2}{2 \, \alpha_1 \, \alpha_2\,  \sum_i G_{i,1} \, G_{i,2}  }$  into  the M-step equation 
 \ref{eq:optimization-criteria}
 gives us the objective function 
 
\begin{eqnarray} 
  O(\{\alpha_k,\theta_k\})=\sum_i \sum_k Q_i(k)\, \log P(p_i,k|\{\alpha_k, \theta_k\})
  \label{eq:objective-function}
 \end{eqnarray}
 and so 
 \begin{eqnarray} 
(\alpha_k^{t},\theta_k^{t})& = &  \arg \max_{\{\alpha_k,\theta_k\}} \sum_i  \left \{  Q_i^{t}(1) \, \log \left [ 2\, \alpha_{1}^2 \,  G_{i,1}^2 
 +  \, (\alpha\, o)_i  \right ]  +\,  Q_i^{t}(2) \, \log \left [ 2\, 
\alpha_{2}^2 \,  \, G_{i,2}^2 
 +  \, (\alpha\, o)_i \right ] \right \} \qquad 
\label{eq:optimization-criteria-Two-classes-nocos}
\end{eqnarray}
 where in the maximization we must satisfy the constraint 
 \begin{eqnarray}
 -1\le \cos \phi=\frac{1-\alpha_1^2 - \alpha_2^2}{2 \, \alpha_1 \, \alpha_2\,  \sum_i G_{i,1} \, G_{i,2}  } \le 1\, .
 \label{eq:constraint-cosine}
\end{eqnarray}
Define the vectors 
 \begin{eqnarray}
 \vec \alpha_{k}= \alpha_k \, \hat g_k \quad {\rm and}  \quad
 (\hat g_k) = (G_{1,k}, G_{2,k}, \hdots, G_{i,k}, \hdots, G_{N,k})^T
 \end{eqnarray}
 where $
 \hat  g_k  \cdot \hat g_k=\sum_{i=1}^N G_{i,k} \, G_{i,k}=\sum_{i=1}^N G_{i,k}^2=1$. 
 
 Thus, $\vec \alpha_{k}\cdot \vec \alpha_{k}=\alpha_k^2$ and 
 $\vec \alpha_{1}\cdot \vec \alpha_{2}=\alpha_1 \, \alpha_2\, \cos \varphi_{1,2}= \alpha_1 \, \alpha_2\,  \sum_{i=1}^N G_{i,1} \, G_{i,2} $. Therefore the  constraint \ref{eq:constraint-cosine} can be written  as
 
 \begin{eqnarray}
\min \{(\vec \alpha_1-\vec \alpha_2)^2, \,  (\vec \alpha_1+\vec \alpha_2)^2\} \le 1 \le \max\{(\vec \alpha_1-\vec \alpha_2)^2,\,  (\vec \alpha_1+\vec \alpha_2)^2\}
 \end{eqnarray}

 \subsection{Gradient Equations for the  M-Step}
 The maximization step can be achieved either at the constraint boundaries or where the gradient with respect to the free parameters are zero.  We now compute these gradients of  the objective function $O(\{\alpha_k,\theta_k\})$ (given in equation \ref{eq:objective-function}) with respect to each of its parameters $(\alpha_k,\theta_k)$ and investigate when the gradient vanishes. Some of the detail calculations are presented in Appendix~\ref{sec:gradient}

\begin{eqnarray}
\mu_k^t &\rightarrow & \partial_{\mu_{k'}} O(\{\alpha_k,\theta_k\}) = 
 \sum_i  \sum_k Q_i^{t}(k) \, \frac{\partial_{\mu_{k'}} P( p_i, k|\{ \theta_k^{t-1} \})}{P( p_i, k|\{ \theta_k^{t-1} \})}
 =0   
 \nonumber \\
{\bf C}_k^{t}&\rightarrow& \partial_{{\bf C}_{k'}} O(\{\alpha_k,\theta_k\})= \sum_i  \sum_k Q_i^{t}(k) \, \frac{\partial_{{\bf C}_{k'}} P( p_i, k|\{ \theta_k^{t-1} \})}{P( p_i, k|\{ \theta_k^{t-1} \})}  =0  
\nonumber \\
\alpha_k
&\rightarrow & \partial_{\alpha_{k'}} O(\{\alpha_k,\theta_k\})=  \sum_i  \sum_{k} Q_i^{t}(k) \, \frac{\partial_{\alpha_{k'}} P( p_i, k|\{ \theta_k^{t-1} \})}{P( p_i, k|\{ \theta_k^{t-1} \})}  =0
\label{eq:M-step} 
\end{eqnarray}

$\mu_{k}$: derived in \ref{eq:mu-final} and \ref{eq:mu-def} 

 \begin{eqnarray}
 \mu_k = \frac{\sum_{i} F_{i,k} \, \, p_{i}}{\sum_{j=1}^N F_{i,k}}
 \label{eq:copy-mu}
 \end{eqnarray}
where  $
F_{i,k} =    Q_i(k)  \, - \, o_i
 \sum_{j=1}^N \frac {\, \frac {1}{2} \,(\alpha\, o)_i }{\alpha_{1}^2 \, G_{j,1}^2 + \alpha_{2}^2 \, G_{j,1}^2 + \frac{1}{2}(\alpha\, o)_j } 
 $
 \\
 
 ${\bf C_{k}}$: derived in \ref{eq:covariance} and \ref{eq:covar-def}

\begin{eqnarray}
{\bf C}_{k} &=& \frac{\sum_i R_{i,k}^t \,  (p_i -\mu_k)\, (p_i-\mu_k)^T}{\sum_i R_{i,k}^t}
\label{eq:copy-Cov}
\end{eqnarray}
where 
$
R_{i,k}^t =  F_{i,k}+ \frac{ \alpha_{k}^2 \, G_{i,k}^2  }{\alpha_{1}^2 \, G_{i,1}^2 + \alpha_{2}^2 \, G_{i,1}^2 + \frac{1}{2}(\alpha\, o)_i} 
 $

$\alpha_{k}$:
\begin{eqnarray}
 {\rm for} & \partial_{\alpha_{1}}& {\rm derived \, in \, } \ref{eq:alpha1}
\nonumber \\ 
0 &=&  \sum_{i}^{N}Q_{i,1}^{t} \,  \frac{  \left ( 2 \,   G_{i,1}^2 \,  -\,o_i\right ) }{\left [ 2\, \alpha_{1}^2 \,  G_{i,1}^2 
 +  \,  (\alpha\, o)_i \right ]} 
 -Q_{i,2}^{t}\,  \frac{  o_i }{\left [ 2\, \alpha{2}^2 \,  G_{i,2}^2 
 +  \,  (\alpha\, o)_i \right ]} 
 \nonumber \\ 
 && {\rm applying\, equation } \ref{eq:E-quantum-step-two-nocos}
 \nonumber \\
 &=& \sum_{i}^{N}  \frac{  \left ( G_{i,1}^2 \,  -\,o_i\right ) }{\left [ \alpha_{1}^2 \,  G_{i,1}^2 
 +  \alpha_{2}^2 \,  G_{i,2}^2 
 + \, (\alpha\, o)_i  \right ]} 
\label{eq:copy-alpha1} \\ 
\nonumber \\ 
 {\rm for} & \partial_{\alpha_{2}}& 
 {\rm derived \, in \, } \ref{eq:alpha2}
\nonumber \\ 
0 &= &    \sum_{i}^{N} -\, Q_{i,1}^{t} \,  \frac{  o_i  }{\left [ 2\, \alpha_{1}^2 \,  G_{i,1}^2 
 +  \,  (\alpha\, o)_i \right ]} 
 + Q_{i,2}^{t}\,\frac{   \left (  2 \,   G_{i,2}^2 \,- o_i\right ) }{\left [ 2\, \alpha_{2}^2 \,  G_{i,2}^2 
 +  \,  (\alpha\, o)_i  \right ]} 
 \nonumber \\ 
 && {\rm applying\, equation } \ref{eq:E-quantum-step-two-nocos}
 \nonumber \\
 &=& \sum_{i}^{N}  \frac{  \left ( G_{i,2}^2 \,  -\,o_i\right ) }{\left [ \alpha_{1}^2 \,  G_{i,1}^2 
 +  \alpha_{2}^2 \,  G_{i,2}^2 
 + \, (\alpha\, o)_i  \right ]} 
 \label{eq:copy-alpha2}
\end{eqnarray}
where $o_i$ and $(\alpha \, o )_i$ are measures of overlap given by equations \ref{eq:overlap-measures}

\subsection{The Classical Limit}
For the case $\phi = \frac{\pi}{2}$, the probabilities \eqref{eq:pik-two-classes} becomes
\begin{eqnarray}
 \begin{matrix}
P(p_i,1|\{\alpha_k, \theta_k\}) &=&
\alpha_{1}^2 \,  \, G_{i,1}^2 
 \\
 P(p_i,2|\{\alpha_k, \theta_k\}) &=&
\alpha_{2}^2 \,  \, G_{i,2}^2 
\end{matrix}
\label{eq:pik-two-classes-classic}
\end{eqnarray}
and $\alpha_1^2+\alpha_2^2=1\, \rightarrow \, (\alpha \, o )_i=0$, \,$ 
F_{i,k} =    Q_i(k) \, \rightarrow \, \mu_k=\frac{ \sum_i Q_{i,k}\, p_i}{ \sum_i Q_{i,k}}
 $,\,  and \,  $R_{i,k}^t=2\, Q_{i,k} \, \rightarrow \, {\bf C_k} =\frac{ \sum_i Q_{i,k}\, (p_i-\mu_k)(p_i-\mu_k)^{T}}{ \sum_i Q_{i,k}}$. We recover the classical mixture model.  The two class case allows us  to compare the performance of the two models, the quantum and the classical.

\section{Experiments and Analysis}
\label{sec:experiments}
We restrict the analysis to the two class cases. The main reason is to compare the performance (empirically as well as theoretically)  to the classical case. 
We first devise experiments with 2D randomly controlled data sets to fully analyze the statistical properties of the quantum inspired method vs the classical mixture of Gaussians (state of the art EM method in clustering). We then devise color segmentation experiments with 3D data  (see section~\ref{sec:color-3D-experiments}).   to illustrate how  the technique can be applied and outperform state of the art techniques in a computer vision application.

\subsection{Statistical Analysis of Controlled Data Sets}
\label{sec:analysis-controlled-dats}

We generated random data $p_i^k =(x_i,y_i)$ with mean   $\mu^{k}=\begin{bmatrix} \mu^{k}_{x} & \mu^{k}_{y}\end{bmatrix}^T$ and  covariance  ${\bf C}=\begin{bmatrix} \cos \theta & \sin \theta \\ -\sin \theta & \cos \theta \end{bmatrix}\begin{bmatrix} (\sigma_{x}^2)^k & 0 \\ 0& (\sigma_{y}^2)^k \end{bmatrix}\begin{bmatrix} \cos \theta & -\sin \theta \\ \sin \theta & \cos \theta \end{bmatrix}$, where $\theta$ is the rotation of the x-axis where    $\sigma_x^k$ is the standard deviation along $x$ for class $k$. In all the tests we considered $N_1=500$ and $N_2=500 \, {\rm or}\, 1000$ points per class. We refer to these parameters $\mu$'s, $\sigma$'s, $N$'s as the ground truth of the experiments, as the samples come from single Gaussians with these parameters and with that many samples. This is  to say that we did not sample from the mixture of Gaussians model. The goal of each experiment was to compare the performance of the quantum model and the classical one, by estimating the number of points of each class, the center of each class and the eigenvalues of each covariance (the variances). The estimations allow us to understand which model best identifies the original Gaussian distribution parameters and number of points (we refer to these parameters as the ground truth).
  \\
 To account for the random process,  50 trials were produced for each combination of variables. The estimated mean  per trial $i=1, ...,50$ is represented as $\begin{bmatrix}\overline{\mu}^k_{x, i}, \overline{\mu}^k_{y, i}\end{bmatrix}^T$, where
 \begin{eqnarray}
 \overline{\mu}^k_{x}
 =\frac{1}{50}\sum_{i=1}^{50} \overline{\mu}^k_{x, i} 
 \end{eqnarray}
In order to compare the estimated mean per trial to the true value of the mean [$\mu^{k*}_{x}$, $\mu^{k*}_{y}$], we show the error, calculated as: 
\begin{eqnarray}
Error(\mu^k_x) \, = \,\sqrt{(\mu^{k*}_{x} - \overline{\mu^k_{x}})^2}
\label{eq:error}
\end{eqnarray}
and we refer to the fluctuation of this error as 
\begin{eqnarray}
Fluctuation(\mu^k_x) \, = \bar \eta^k_{x}=\,\sqrt{\frac{1}{50}\sum_{i=1}^{50}(\mu^k_{x,i} -\overline{\mu}^k_x )^2}
\label{eq:fluctuation}
\end{eqnarray}
The fluctuation does not depend on the ground truth, as it characterize the variation of the statistical experiments. 

This error and fluctuations were calculated for all the parameters, namely the means (or centers) $\mu^k_{x,y}$ (for x and y separately), variances $(\sigma^k_{x,y})^2$  (for x and y separately), and the number of points per class (N1 and N2).


\subsubsection{Tests}

\begin{figure}[h!]
\begin{minipage}[b]{.55\linewidth}
\centering
\includegraphics[width=\linewidth]{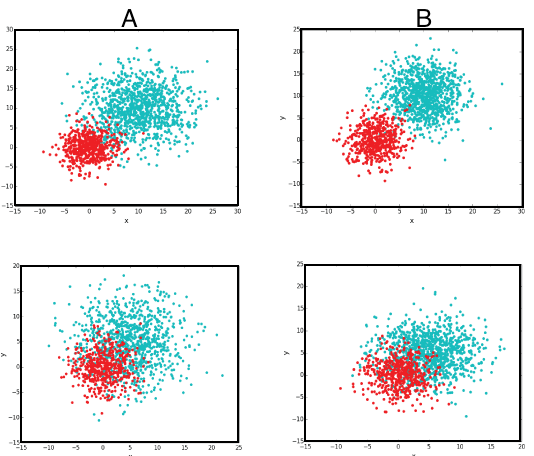}\\
\end{minipage}
\centering
\caption{\small For all four images there are 500 red dots in Class 1 (N1), and 1000 blue-green dots in Class 2 (N2). Also, for all four images $\mu^1_{x,y}$ = (0,0) and $\sigma^1_{x,y}$ = 3. First column, A:  $\sigma^2_{x,y}$ = 5. Second column, B:    $\sigma^2_{x,y}$ = 4.  Top row:  'separated' distributions with $\mu^2_{x,y}$ = 10, and Bottom row: overlapping distributions with $\mu^2_{x,y}$ = 5.}
\label{fig:colored-data-sets}
\end{figure}

We arbitrarily fixed $N^1=500$ and $N^2=1000$ to generate the data sets, so the sizes of each class are different and large enough for statistical measurement.  The Gaussian parameters to generate the  $N^1=500$ data points for  the first class were $\mu^1_{x,y}$ = 0, $\sigma^1_{x,y}$ = 3. 
The center of the second Gaussian distribution is varied starting with $\mu^2_{x,y} =10$ and gradually brought closer together to $\mu^1_{x,y}$ = 0 to create greater overlap. As we brought  the second center towards the first center, we repeated experiments for $\sigma^2_{x,y}$ = 4 and 5  (see some examples of data sets in  figure~\ref{fig:colored-data-sets}.)

\begin{figure}[h!]
\begin{minipage}[b]{1.00\linewidth}
\centering
\includegraphics[width=\linewidth]{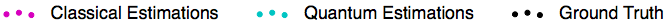}\\
\end{minipage}
\begin{minipage}[b]{.75\linewidth}
\centering
\includegraphics[width=\linewidth]{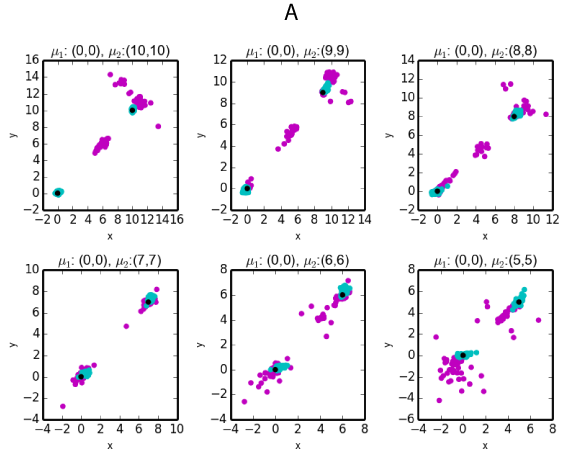}\\
\end{minipage}
\begin{minipage}[b]{.75\linewidth}
\vspace{.5cm}
\centering
\includegraphics[width=\linewidth]{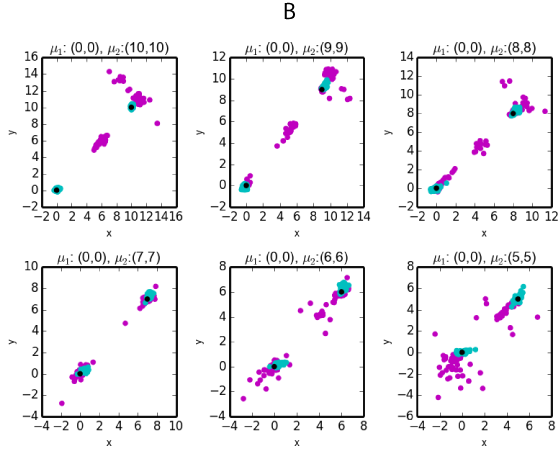}\\
\end{minipage}
\centering
\caption{\small  Each dot represents a center estimation for each trial. Purple dots represent estimation for the classical method while Cyan dots represents estimation from the Quantum method. The two Black dots per image represent the ground truth centers.  For all 12 images , $\mu^1_{x,y}$ = 0,  $\sigma^1_{x,y}$ =  3,  N1 = 500 and N2 = 1000. A shows $\sigma^2_{x,y}$ = 5 and B shows $\sigma^2_{x,y}$ = 4.  Both, A and B, show results as $\mu^2_{x,y}$ vary from 10 to 5. It is  noticeable how the trials for the quantum method almost always provide good answers (and it is robust), while the classical method has less accuracy and much more variation. We quantify this assertion in figure~\ref{fig:variance-estimation-errors}} 
\label{fig:centers-visualization}
\end{figure}

\begin{figure}[h!]
\centering
\begin{minipage}[b]{.23\linewidth}
\includegraphics[width=\linewidth]
{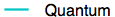}
\end{minipage}
\begin{minipage}[b]{.23\linewidth}
\includegraphics[width=\linewidth]
{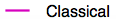}
\end{minipage}
\centering
\begin{minipage}[b]{1.0\linewidth}
\centering
\includegraphics[width=\linewidth]{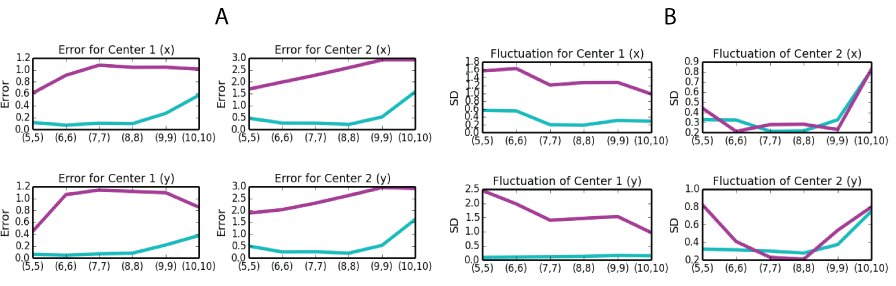}
\end{minipage}
\textbf{Results for centers} \\
\centering
\begin{minipage}[b]{1.0\linewidth}
\centering
\includegraphics[width=\linewidth]{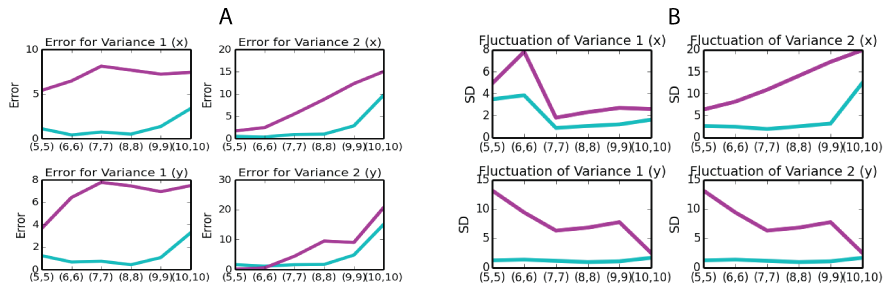}
\end{minipage} 
\textbf{Results for variances}\\
\centering\caption{ Y-axis: Column A.) Estimated error (given by equation \ref{eq:error}) and Column B.) standard deviation (fluctuations)(given by equation \ref{eq:fluctuation}). Top two rows are   for both centers and x and y coordinates. Bottom two rows are  for both variances and x and y coordinates. The X-axis for all graphs: varying $\mu^2_{x,y}$ values from $(5,5)$ to $(10,10)$. In these trials, the true $\mu^1_{x,y} = 0$, and variances, $(\sigma^1_{x,y})^2 = 9$ and $(\sigma^2_{x,y})^2$ = 25. The quantum method proves to be more accurate and robust, as it gives a lower error and fluctuates less. As $\mu^2_{x,y}$ increases, there is less overlap between the two distributions. In the overlapping regimes, the highly fluctuating results of the classical case indicate that the method is unable to identify the parameters of the separate groups. We see the effects of this on the error curve, as the average of highly fluctuating estimated variables can be closer the ground truth. }
\label{fig:variance-estimation-errors}
\end{figure}

\begin{figure}[h!]
\centering
\begin{minipage}[b]{.20\linewidth}
\includegraphics[width=\linewidth]
{Figures/Quantum.png}
\end{minipage}
\begin{minipage}[b]{.20\linewidth}
\includegraphics[width=\linewidth]
{Figures/Classical.png}
\end{minipage}
\begin{minipage}[b]{1.00\linewidth}
\includegraphics[width=\linewidth]
{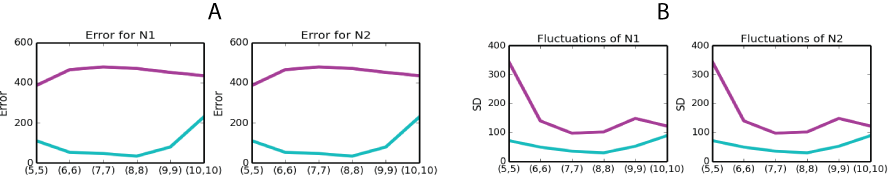}
\end{minipage}
\textbf{Results for number of points} \\
\caption{\small Y-axis: Estimated error for $N^1$, $N^2$ calculated using equation~\ref{eq:error}  and standard deviation calculated using equation~ \ref{eq:fluctuation} vs   X-axis: $\mu^2_{x,y}$ is varied. Because the quantum method returns a lower error and standard deviation (fluctuations), it is the more accurate and robust method. As the data is more separated, the difference between the methods is reduced. Estimated error values for N1 and N2 are  identical since $N^1+N^2=N$ is known/constant.  }
\label{fig:Number-of-points-per-class}
\end{figure}

For each experiment there were 50 trials (see figure~\ref{fig:centers-visualization}). We then  plot the errors (given by equation~ref{eq:error} and fluctuations (given by equation~\ref{eq:fluctuation}) for each experiment (see figures ~\ref{fig:variance-estimation-errors} and \ref{fig:Number-of-points-per-class}). 

\clearpage
\subsubsection{Overlap and Interference}
\begin{figure}[h!]
\begin{center}
\begin{minipage}[b]{0.50\linewidth}
\includegraphics[width=\linewidth]
{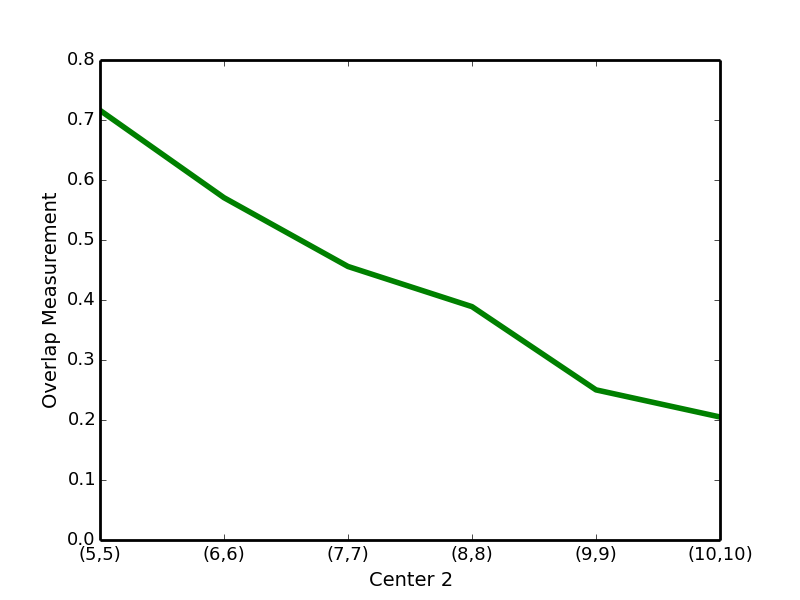}
\centering
\textbf{Overlap as distributions separate}\\
\end{minipage}
\scalebox{0.85}{\vspace{.7cm}
\begin{tabular}{l |l|l| l | l| l|l }
\hline
Overlap & 0.716 & .571 & 0.456 & 0.389 & 0.250 & .205\\
\hline
$\cos\phi$ & 3.59E-6 & 8.68E-6 & 7.01E-5 & 5.24E-5&5.93E-5& 0.173\\
\end{tabular}}
\end{center}
\caption{\small Y-axis: Overlap $\sum_i \, G_{i,1} \, G_{i,2}$, as $\mu^2_{x,y}$ varies on the X-axis for the fixed parameters  $\mu^1_{x,y}$ = 3, $\sigma^1_{x,y}$ =3, $\sigma^2_{x,y}$ = 5, $N^1$ = 500 and $N^2$ = 1000.  When $\mu^2_{x,y}$ = 10, we see that there is still some overlap, measured approximately to be .205. This sheds light on our results in Figure 4, where the  classical model still under performs compared to the quantum model. The classical method is highly sensitive to any overlaps while the quantum model chooses a phase difference between classes to accurately recover the Gaussian parameters and number of points per class. This is shown by the table above which calculates $\cos\phi$ experimentally using $\cos \phi=\frac{1-\alpha_1^2 - \alpha_2^2}{2 \, \alpha_1 \, \alpha_2\,  \sum_i G_{i,1} \, G_{i,2}  }$}.
\label{fig:overlap}
\end{figure}

We argue that  the quantum method utilizes the phase parameter to help capture the overlapping of the two distributions. In the quantum method the phase differences between classes yields the interference phenomena, which is stronger as the data overlaps. We want to demonstrate that the interference better captures the overlapped data. For this, through out the experiments described above, we measure the overlap between the data as shown in figure~\ref{fig:overlap}  and we show the phase difference (measured by the cosine of it in the table of figure~\ref{fig:overlap}). 
\subsection {Assessment of Log Likelihood Landscapes}
To better understand why the quantum method outperforms the classical method, we first assess the landscapes  objective function, $ O(\alpha_k^{t},\theta_k^{t})=  \sum_{i=1}^N  \sum_{k=1}^2  Q_i^{t}(k) \, \log  P( p_i, k|\{ \theta_k^{t-1} \}) $, to examine how the values obtained fair with the desired solution. Note that the objective function is different for the quantum method (see \ref{eq:joint-distribution-quantum} and \ref{eq:E-quantum-step-two-nocos}) and the classical method (see \ref{eq: classical-distribution} and \ref{eq:E-step}), due to the calculation of $P(p_{i}, k| \{\theta_k\})$ and $Q_{i}$(k) in each method. We wonder if a local minima is reached  which might prevent variables at converging at the correct values. Second, we also asses the final solution's objective functions to see which method offers a better cost function (assessing the interference phenomena ability to lower the cost function). Third, in generating these landscapes, we also test to see whether initial values of the the centers have any affect on the estimated variables. 

In order to create the landscapes we fix all parameters but $\mu^1_x$ and $\alpha_1$ (the parameter $\alpha_2$ varies, but is determined from $\alpha_1$). The sample over the $\mu^1_x$ values is in the range between the minimum and maximum  values from the EM iterations, and in increments of 0.01. For the classical method we do the same, we get the maximum and minimum values of $\alpha_1$ and sample by increments of 0.01 in between these values. However, for the  quantum method, to fully specify $\alpha_2$ (and the objective function), we need the phase values as well as $\alpha_1$. Thus, for the quantum method, we sample over the $\alpha^1$ values  obtained from each iteration of the EM procedure. The number of iterations is not predetermined, but rather depends when there is convergence.
 
For both the classical and quantum methods, we note the objective function during the initialization of EM, and the during the ending of EM to ensure that the objective function did indeed increase, and is close to the true objective function. For the classical method, the true objective function was obtained by fixing all the parameters in \ref{eq:optimization-criteria}. For the quantum method, all the parameters except $\alpha_1$ and $\alpha_2$ were fixed. In order to obtain those two variables, we performed an exhaustive search over the [0.0, 1.0] in increments of .001, to obtain the retrieve the lowest possible objective function defined by \ref{eq:objective-function}.

\begin{figure}[h!]
\begin{center}
\textbf{Quantum Landscapes}
\begin{minipage}[b]{0.99\linewidth}
\includegraphics[width=\linewidth]
{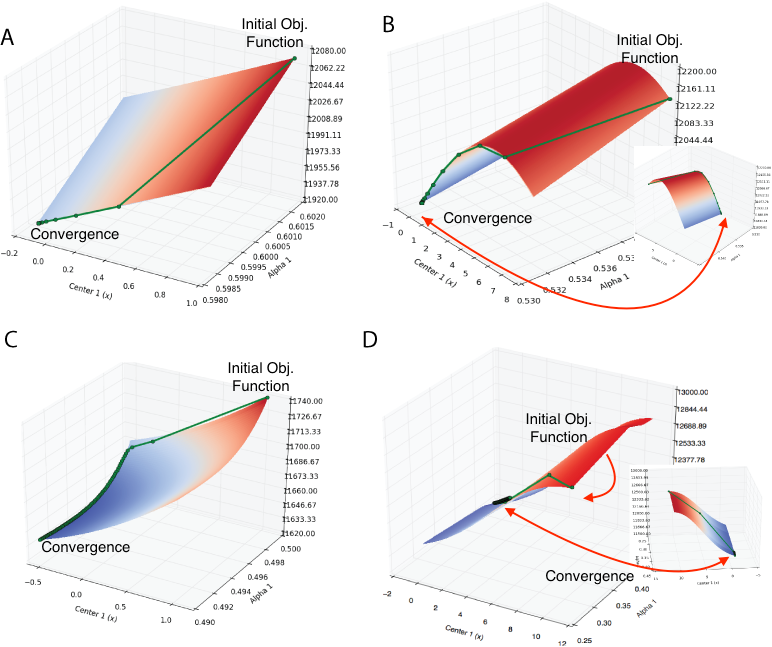}
\end{minipage}
\end{center}
\begin{center}
Estimated Values for Quantum Method
\scalebox{0.85}{
\begin{tabular}{l |l|l| l | l| l|l|l}
\hline 
Image & Initial  & Final  &  Estimated& Estimated & Initial & Final & True \\
&$\mu^1_{x}$ & $\mu^1_{x}$&N1 & N2&Obj. Function& Obj. Function & Obj. Function\\
\hline
A&$1$ &-0.1289 &438.572 &1061.427 &12045.968 &11909.624&11941.578 \\
B&$8$ & -0.287&376.447 &1123.553& 12113.659&11895.729&11941.578\\
C&$1$ &-0.073 &460.572 &1039.023 & 11620.084 &11909.624&11535.664 \\
D&$11$ & -0.568& 456.662 &1043.337& 12632.097&11628.296&11535.664\\
\end{tabular}}\\
\end{center}
\caption{\small These landscapes from the quantum method show the objective function with changing $\mu^1_{x}$ and $\alpha_1$. For A,B, C, and D, we fixed the parameters to  $\mu^1_{y}$ = 0, $\sigma^1_{x,y}$ = 3, $\sigma^2_{x,y}$ = 5, N1 = 500, and N2 = 1000. For cases A and B, $\mu^2$ was fixed to 5, and for cases C and D, $\mu^2$ was fixed to 9. The table above specifies the original/true objective function, which is lower than the ones for the classical case, due to the interference phenomena.}
\end{figure}
\begin{figure}[h!]
\begin{center}
\textbf{Classical Landscapes}
\begin{minipage}[b]{0.99\linewidth}
\includegraphics[width=\linewidth]
{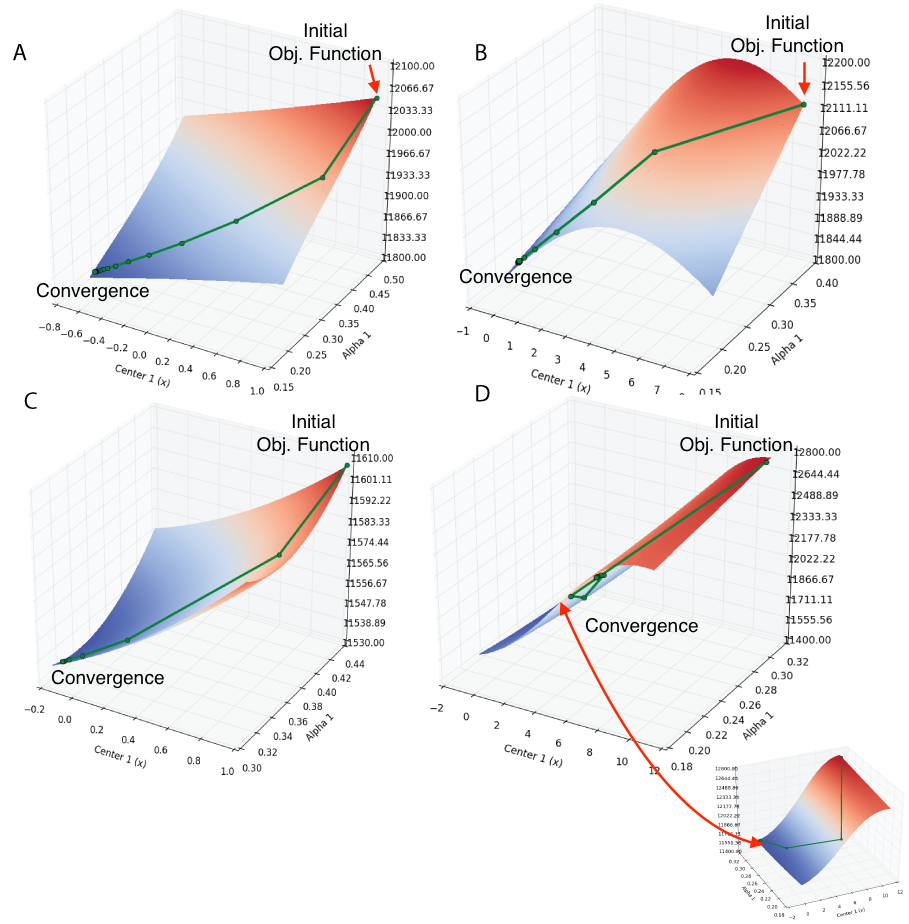}
\end{minipage}
\end{center}
\begin{center}
Estimated Values for Classical Method
\scalebox{0.85}{
\begin{tabular}{l|l |l| l | l| l|l|l}
\hline 
Image &Initial  & Final  &  Estimated& Estimated  &Initial & Final & True \\
&$\mu^1_{x} $ & $\mu^1_{x}$ &N1 & N2&Obj. Function& Obj. Function & Obj. Function\\
\hline
A&$1$ &-0.551 &326.469 &1173.531 &12047.035 &11825.221&12176.206\\
B&$8$ & -0.845&306.459 &1193.540& 12113.659&11814.813&12176.206\\
C&$1$ &-0.279 &462.194 &1037.805 & 11586.998 &11570.7837&11577.100 \\
D&$11$ &-0.120& 463.622&1036.377& 12476.823&11539.203&11577.100\\
\end{tabular}}\\
\end{center}
\caption{\small These landscapes from the classical method show the objective function with changing $\mu^1_{x}$ and $\alpha_1$. For A,B, C, and D, we fixed the parameters to  $\mu^1_{y}$ = 0, $\sigma^1_{x,y}$ = 3, $\sigma^2_{x,y}$ = (5,5), N1 = 500, and N2 = 1000. For cases A and B, $\mu^2$ was fixed to 5, and for cases C and D, $\mu^2$ was fixed to 9. The table above specifies the original/true objective function, which is higher than the ones for the quantum case, due to the interference phenomena. The objective function for the final values is lower than the original/true objective function, which means the method is performing adequately. }
\end{figure}
\clearpage
\subsection{Deformations to the Gaussian Assumption}
\label{sec:Deformations-Gaussians}
So far, the tested classes were generated from  Gaussian distributions. Now, in order to deform the shape of the distribution, the data set is initially drawn from a Gaussian distribution and then, each  point coordinates is displaced by  a random number drawn from a uniform distribution in $[-\epsilon,\epsilon]$.  
In order to remove any biases that would cause the quantum method to outperform its classical counter part, we start with distributions where both methods performed equally well. This was the case for which the distributions were separated with an overlap measurement of 6.82E-12, and both classes were equal in covariance and in the number of points.  The parameters for both classes were fixed at $\mu^1_{x,y} = 0 $, $\mu^2_{x,y} = 7$, and variances $(\sigma^1_{x,y})^2 \, = \,  (\sigma^2_{x,y})^2 \,  = \, 9$, $N^1 = N^2 $= 500. 50 trials were run for $\epsilon$ values of 0.00, 0.75, 1.50, 3.00, 3.75, 4.50, 5.25, and 6.00. (see figure 7, 8, 9).
The quantum  method outperforms the classical method, proving to have a more robust performance to distributions that are not "ideal Gaussian distributions". 
\begin{figure}[h!]
\begin{minipage}[b]{.20\linewidth}
\includegraphics[width=\linewidth]
{Figures/Quantum.png}
\end{minipage}
\begin{minipage}[b]{.20\linewidth}
\includegraphics[width=\linewidth]
{Figures/Classical.png}
\end{minipage}
\vspace{.7cm}
\centering
\includegraphics[width=\linewidth]{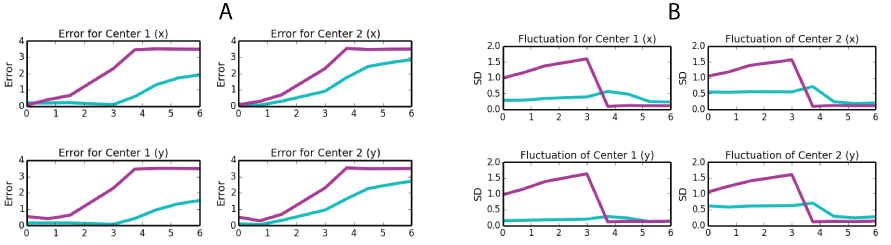}
\textbf{Results for centers}\\
\caption{\small Y-axis: Estimated error (given by equation \ref{eq:error}) and standard deviation (fluctuations)(given by equation \ref{eq:fluctuation})  for both centers vs X-axis:  the deformation $\epsilon$ was increased. Throughout these experiments, to avoid any advantage to the quantum method, we drew the data from  $\mu^2_{1,2}$ = $\mu^2_{1,2}$ = 3, and variances $(\sigma^1_{x,y})^2$ =  $(\sigma^2_{x,y})^2$ = 9, and $N^1$ = $N^2$ = 500. Without the deformations, both methods perform similarly, as evidenced by results at $\epsilon$ = 0.}
\label{fig:center-deformation-data}

\begin{minipage}[b]{1.0\linewidth}
\centering
\includegraphics[width=\linewidth]{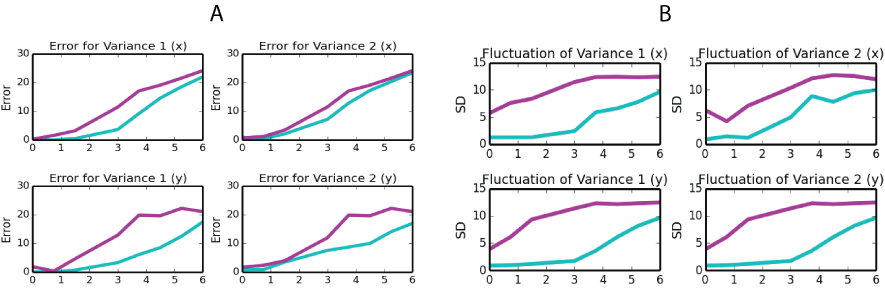}\\
\centering
\textbf{Results for variances}\\
\caption{\small Y-axis: Estimated error (given by equation \ref{eq:error}) and standard deviation (fluctuations) (given by equation \ref{eq:fluctuation}) for both variances $(\sigma^1_{x,y})^2$ and $(\sigma^2_{x,y})^2$ vs  X-axis: increasing  $\epsilon, \mu^2_{x,y}$. In these trials, the true $(\sigma^1_{x,y})^2 = 9$ and $(\sigma^2_{x,y})^2$ = 25. The quantum method outperforms on both accuracy and robustness since it returns a lower error on the average variances and fluctuates less.}
\end{minipage}
\label{fig:variance-deformation-data}
\end{figure}

\clearpage
\begin{center}
\begin{figure}[h!]
\centering
\begin{minipage}[b]{.20\linewidth}
\includegraphics[width=\linewidth]
{Figures/Quantum.png}
\end{minipage}
\begin{minipage}[b]{.20\linewidth}
\includegraphics[width=\linewidth]
{Figures/Classical.png}
\end{minipage}\\
\begin{minipage}[b]{1.00\linewidth}
\centering
\includegraphics[width=\linewidth]{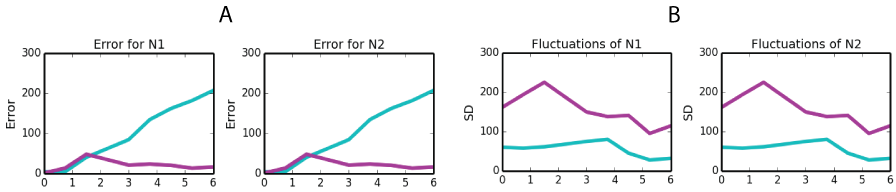}
\end{minipage}
\textbf{Results for number of points}\\
\label{fig:number-points-deformation-data}
\caption{\small Y-axis: Estimated error (given by equation \ref{eq:error}) and standard deviation (fluctuations)(given by equation \ref{eq:fluctuation})  for N1 (first column) and N2 (second column) vs X-axis:  the deformation $\epsilon$ was increased.   Notice that after the classical system has completely failed to identify the other parameters accurately, the standard deviation decreases as does the error for identifying $N^1$ and $N^2$. This suggests a bias towards $N^1=N^2$, even if estimations are wrong.}
\end{figure}
\end{center}
\subsection{ Experiments in Color Segmentation: 3D Data}
\label{sec:color-3D-experiments}
Color images have each pixel represented in RGB space, i.e., for each pixel three numbers are assigned representing the amount of red (R), green (G), and blue (B). The color segmentation problem consists of assigning  to each pixel one class, foreground or background (1 or 0). From a clustering point of view, each pixel is one color data point, in 3D (in RGB space), and the task is to discover the two classes in color space and assign each pixel to a corresponding class.  
\begin{figure}[h!]
\begin{center}
\begin{minipage}[b]{.20\linewidth}
\includegraphics[width=\linewidth]{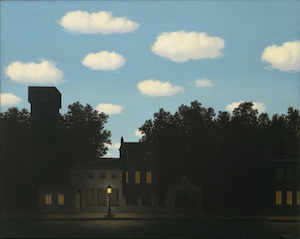}\\{\small (a) RGB Image}
\end{minipage} 
\begin{minipage}[b]{.20\linewidth}
\includegraphics[width=\linewidth]{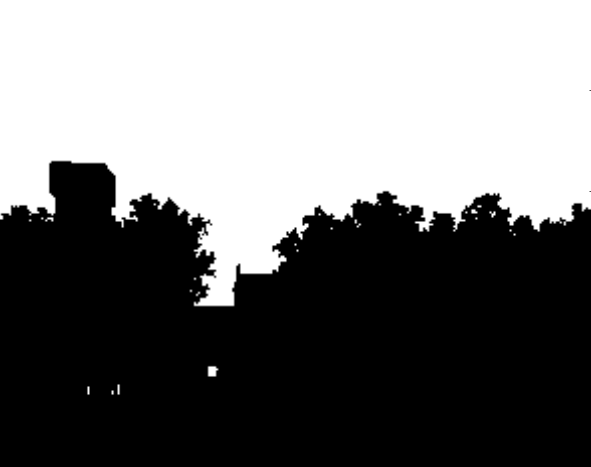}\\{\small (b) Binary Image}
\end{minipage}
\caption{The RGB image of Magritte's \textit{Empire of Light}. Each pixel was assigned either 0 or 1 (for black or white) by creating a threshold for each RGB channel. If red, green and blue levels in the original image were all above 100, then the pixel was assigned to 1, and 0 otherwise.}
\end{center}
\label{fig:RGB and Binary Image}
\end{figure}

Of course, in computer vision, one does employ the notion of neighborhood, that nearest pixels are more likely to belong to the same class. Here, we only focus on the clustering problem, i.e., each pixel is treated as independent. To build a complete color segmentation problem, one could use the final clustering probabilities to a "neighborhood" prior and work on the posterior probability. This would lead to a graph partitioning problem that is known to be solved with the min cut algorithm \cite{Ishikawa1998, Boykov2001}. However,  our goal here is to focus on data probability (likelihood) problem, i.e., on the clustering probabilities. To demonstrate that our quantum inspired method works significantly better than the state-of-the-art Gaussian mixture model.

\paragraph{L*a*b* color space:} A Lab color space is a color-opponent space with dimensions L for lightness and a and b for the color-opponent dimensions, based on non-linearly compressed  coordinates (e.g. CIE XYZ) (\cite{K1976}). The lightness, L*, represents the darkest black at L* = 0, and the brightest white at L* = 100. The color channels, a* and b*, will represent true neutral gray values at a* = 0 and b* = 0. The red/green opponent colors are represented along the a* axis, with green at negative a* values and red at positive a* values. The yellow/blue opponent colors are represented along the b* axis, with blue at negative b* values and yellow at positive b* values. It has been used in early computer vision color work precisely because it has good perceptual metric properties (e.g., \cite{Tomasi98, K1976}).  

\paragraph{Setting up the Clustering Problem in 3D L*a*b* space:}
We first colored binary images (black and white) according to  two corresponding class Gaussian distributions, $G^b$ and $G^w$ generated in L*a*b* color space. 

The experiment is then defined as follows. Each image pixel is mapped to a 3D point coordinate $(L^k_i,\, a^k_i, \,b^k_i)$. Then, we extract  $\mu^k_{l,a,b}$=$(L^k, a^k, b^k)$ and the covariance, ${\bf C^k}=\begin{bmatrix} \sigma^k_{L} & 0 & 0 \\ 0& \sigma^k_{a}\ & 0 \\ 0 & 0& \sigma^k_{b} \, \end{bmatrix}$ from the set of points (as many as image pixels).  Once these parameters are extracted (either from the Classical Gaussian of Mixture method or by the Quantum inspired method) we  assign a class to each pixel. 

\paragraph{Class Assignment:} According to the clustering method (Classical or Quantum) the 
  final expectation values $Q^b_{i}; i=1,...,N$, and $Q^w_{i}; i=1,...,N$ determine the class, where $b$ indicates the class we assign the color "black" and $w$ indicates the other class, that we assign the color "white". These distributions were used to compute the number of points per class by both methods.  Thus,  in order to produce the final segmentation, given a pixel, $p_i; i=1, ...,N$, if $Q^b_{i}-Q^w_{i} > \, 0$, then we assigned the  "black" color, and otherwise, we assigned "white" color.  
 
 \paragraph{Results on Artificially Colored Images:}
For each clustering method, classical and quantum inspired, we ran 10 trials for each image, starting each time at a random initialization. To account for the variability in the classical case, we used the image which had estimated parameters closest to the average of its cohort.
\\To quantify the comparison of the methods, we calculated the estimated error for the variables $\mu^k_{L,a,b}$, $\sigma^{k}_{L,a,b}$, and $N^{k}$. 

We found a combined error using {\small 
\begin{eqnarray}
Error=\sqrt{(\mu^{k*}_{L} - \overline{mu^{k}_L})^2 + (\mu^{k*}_{a} - \overline{mu^{k}_a})^2 + (\mu^{k*}_{b} - \overline{mu^{k}_b})^2}\, ,
\label{eq:error-color}
\end{eqnarray}}
where $\mu^{k*}$ denotes the ground truth, and $\overline{mu^{k}_L}$, the estimated value.
\begin{figure}[h!]
\begin{center}
\begin{minipage}[b]{.15\linewidth}
\includegraphics[width=\linewidth]{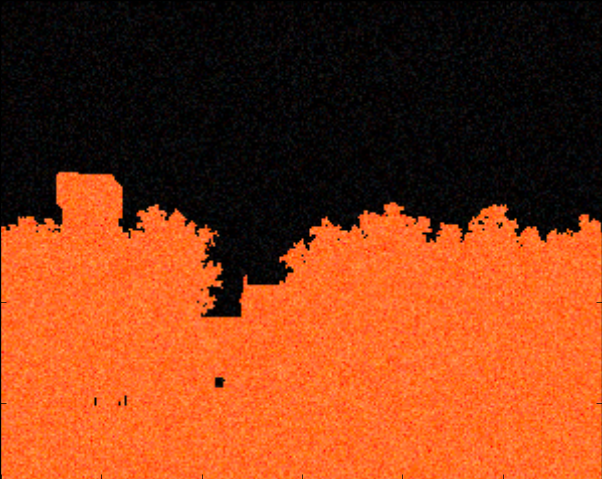}\\{\small Manipulated RGB Image}
\end{minipage}
\begin{minipage}[b]{.15\linewidth}
\centering
\includegraphics[width=\linewidth]{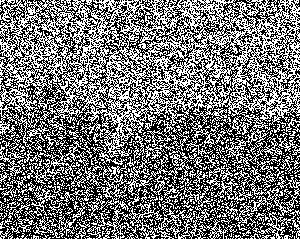}\\{\small Classical Segmentation}
\end{minipage}
\begin{minipage}[b]{.15\linewidth}
\centering
\includegraphics[width=\linewidth]{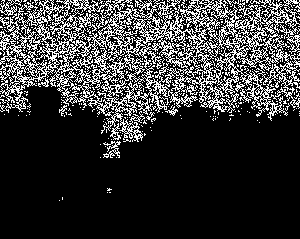}\\{\small Quantum Segmentation}
\end{minipage}\\
\end{center}
\caption{Every pixel in binary image seen in figure 10
was assigned a random point in L*a*b* space generated from  corresponding class Gaussian distributions. The fixed parameters for this case, which defined the class Gaussian distributions were $\mu^w_{L,a,b}$ = (0,0,0), $\mu^b_{L,a,b}$ = (75,75,75), $\sigma^w_{L,a,b}$ = (4,4,4), $\sigma^b_{L,a,b}$ = (6,6,6), $N^w$ = 34034 pixels, and $N^b$ = 37666 pixels.}
{\small The table below summarizes the average of quantum and classical estimated errors using \ref{eq:error-color} and relative error is the ratio of the Quantum error to the Classical one.} \\
\begin{center}
\scalebox{0.75}{
\begin{tabular}{l |l| l | l}
\hline 
Variables & Classical & Quantum & Relative\\
&Error & Error & Error\\
\hline
$\mu^w_{L,a,b}$ & 66.579 & 0.246  & 0.003\\
$\mu^b_{L,a,b}$ & 60.079  & 44.410 & 0.739\\
$N^w$& 1015.121& 19598.009 & 19.306\\
$N^b$&1015.121& 19598.009& 19.306\\
$(\sigma^w_{L,a,b})^2$& 64.721 & 61.488 & 0.950\\
$(\sigma^b_{L,a,b})^2$& 65.023 & 5.567 & 0.085\\
\end{tabular}}\\
\end{center}
{\small Note that although the error in number of point estimation of the classical case is reasonable, this is solely because this image has about the same number of points in both classes but the actual assignment to classes is very poor. So while the quantum method did poorly on the "black region" of the manipulated image, it accurately marked the "orange region" while classical performed poorly everywhere.}
\end{figure}
\clearpage
\begin{figure}[h!]
\begin{center}
\begin{minipage}[b]{.15\linewidth}
\includegraphics[width=\linewidth]
{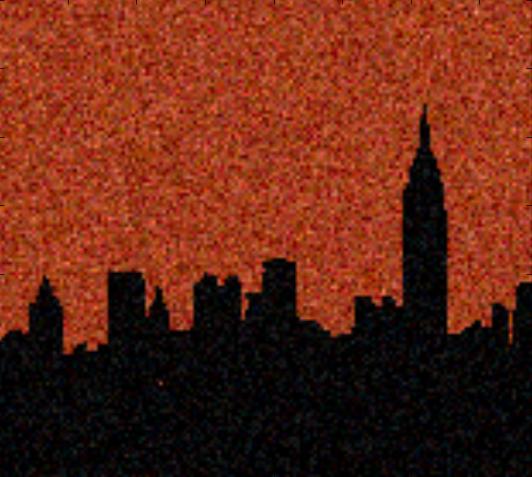}\\{\small Manipulated RGB Image}
\end{minipage}
\begin{minipage}[b]{.15\linewidth}
\centering
\includegraphics[width=\linewidth]{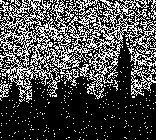}\\{\small Classical Segmentation}
\end{minipage}
\begin{minipage}[b]{.15\linewidth}
\centering
\includegraphics[width=\linewidth]{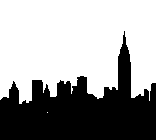}\\{\small Quantum Segmentation}
\end{minipage}
\end{center}
\caption{An original RGB image, available at  
http://alexgoldblum.com/new-york/nyc-skyline/, was binarized. We then colored it artificially randomly sampling each pixel from the  L*a*b*  with the following tow Gaussian  parameters $\mu^w_{L,a,b}$ = (40,40,40),  $\mu^b_{L,a,b}$ = (0,0,0), $\sigma^b_{L,a,b} = (5,5,5)$,  and $\sigma^w_{L,a,b} = (8,8,8)$. The number of pixels in $N^b$ = 8362, and $N^w$ = 13478.}
{\small The table below summarizes the average of quantum and classical estimated errors using \ref{eq:error-color} and relative error is the ratio of the Quantum error to the Classical one.}\\
\begin{center}
\scalebox{0.75}{
\begin{tabular}{l |l| l | l}
\hline 
Variables & Classical & Quantum & Relative\\
& Error & Error & Error\\
\hline
$\mu^w_{L,a,b}$ &42.449& 0.110& 0.002\\
$\mu^b_{L,a,b}$ &26.332& 0.092&0.003\\
$N^w$& 3418.877& 31.000& .009\\
$N^b$& 3418.877 & 31.000& .009\\
$(\sigma^w_{L,a,b})^2$&34.117 &3.948&0.115\\
$(\sigma^b_{L,a,b})^2$&33.466 &1.131&0.034\\
\end{tabular}}
\end{center}
\end{figure}
\section{Conclusion}

In this paper we solely investigated the clustering problem. The state of the art clustering method is EM as it gives close-form solutions for the (Classical) Gaussian Mixture Model (GMM) \cite{Bilmes1998}.  Inspired by quantum methods \cite{Feynman1971} in which interference phenomena causes probability cancellations,  we reformulated this classical model in a quantum model and applied the EM method of parameter estimation.

We showed that the quantum method outperforms the classical (GMM) method in every aspect of the estimations, in cases where the two clusters have an uneven number of data points and different covariances. The Quantum method is able to recover more accurate estimations of all distribution parameters, with much less fluctuations. In regimes where the distributions are separated, we observe that the classical model is very sensitive to slight overlaps. We also show the Quantum method is robust to data deformations from the Gaussian assumptions and is able to segment colors more accurately in the color experiments. 

We plotted the landscapes of each model and showed them to be smooth, leading often to global minima solutions.  There were cases, when the data was very separated and the initial condition very far from the desired solution that the quantum method reached local minima and not global minima. We noted that the quantum method has an extra variable to be computed, the phase of each class.

The biggest difference between the two models is that the cost function for the quantum model  returns a lower value in an overlapping case than the classical solution, suggesting that the interference phenomena drives the quantum system. This allowed the quantum method to produce much more accurate and robust solutions, compared to the classical method. 

For working towards segmentation, clustering may also be interpreted in a Bayesian method as the data likelihood model and  combined with the prior model on the data (such as encouraging neighbors  to be labeled similarly) to form the posterior probability. In this case, for the future,  once the clustering model is produced,   techniques such as graph cuts\cite{Ishikawa1998,Boykov2001} could  be employed  to yield final and better results for color segmentation.

\clearpage
\appendix
\section{Gradient Equations For the Parameters}
\label{sec:gradient}
$\alpha_{k}$:
\begin{eqnarray}
 {\rm for} & \partial_{\alpha_{1}}& 
\nonumber \\ 
0 &=& \sum_{i}^{N}Q_{i,1}^{t} \,  \frac{  \left ( 2 \,   G_{i,1}^2 \,  -\,\frac{G_{i,1} \, G_{i,2}}{ \sum_i G_{i,1} \, G_{i,2}}\right ) }{\left [ 2\, \alpha_{1}^2 \,  G_{i,1}^2 
 +  \,  (1-\alpha_1^2 - \alpha_2^2) \, \frac{G_{i,1} \, G_{i,2} }{ \sum_i G_{i,1} \, G_{i,2}} \right ]} 
 -Q_{i,2}^{t}\,  \frac{  \frac{G_{i,1} \, G_{i,2}}{ \sum_i G_{i,1} \, G_{i,2}}  }{\left [ 2\, \alpha_{2}^2 \,  G_{i,2}^2 
 +  \,  (1-\alpha_1^2 - \alpha_2^2) \, \frac{G_{i,1} \, G_{i,2} }{ \sum_i G_{i,1} \, G_{i,2}} \right ]} 
 \nonumber \\ 
 &=& \sum_{i}^{N}Q_{i,1}^{t} \,  \frac{  \left ( 2 \,   G_{i,1}^2 \,  -\,o_i\right ) }{\left [ 2\, \alpha_{1}^2 \,  G_{i,1}^2 
 +  \,  (\alpha\, o)_i \right ]} 
 -Q_{i,2}^{t}\,  \frac{  o_i }{\left [ 2\, \alpha{2}^2 \,  G_{i,2}^2 
 +  \,  (\alpha\, o)_i \right ]} 
 \nonumber \\ 
 && {\rm applying\, equation } \ref{eq:E-quantum-step-two-nocos}
 \nonumber \\
 &=& \sum_{i}^{N}  \frac{  \left ( G_{i,1}^2 \,  -\,o_i\right ) }{\left [ \alpha_{1}^2 \,  G_{i,1}^2 
 +  \alpha_{2}^2 \,  G_{i,2}^2 
 + \, (\alpha\, o)_i  \right ]} 
\label{eq:alpha1}\\ 
\nonumber \\ 
 {\rm for} & \partial_{\alpha_{2}}&  
\nonumber \\ 
0 &= &  \sum_{i}^{N} -\, Q_{i,1}^{t} \,  \frac{  \frac{ G_{i,1} \, G_{i,2}}{ \sum_i G_{i,1} \, G_{i,2}}  }{\left [ 2\, \alpha_{1}^2 \,  G_{i,1}^2 
 +  \,  (1-\alpha_1^2 - \alpha_2^2) \, \frac{G_{i,1} \, G_{i,2} }{  \sum_i G_{i,1} \, G_{i,2}} \right ]} 
 + Q_{i,2}^{t}\,\frac{   \left (  2 \,   G_{i,2}^2 \,- \frac{G_{i,1} \, G_{i,2}}{ \sum_i G_{i,1} \, G_{i,2}}\right ) }{\left [ 2\, \alpha_{2}^2 \,  G_{i,2}^2 
 +  \,  (1-\alpha_1^2 - \alpha_2^2) \, \frac{G_{i,1} \, G_{i,2} }{  \sum_i G_{i,1} \, G_{i,2}} \right ]} 
 \nonumber
  \\
  &= &  \sum_{i}^{N} -\, Q_{i,1}^{t} \,  \frac{  o_i  }{\left [ 2\, \alpha_{1}^2 \,  G_{i,1}^2 
 +  \,  (\alpha\, o)_i \right ]} 
 + Q_{i,2}^{t}\,\frac{   \left (  2 \,   G_{i,2}^2 \,- o_i\right ) }{\left [ 2\, \alpha_{2}^2 \,  G_{i,2}^2 
 +  \,  (\alpha\, o)_i  \right ]} 
 \nonumber \\ 
 && {\rm applying\, equation } \ref{eq:E-quantum-step-two-nocos}
 \nonumber \\
 &=& \sum_{i}^{N}  \frac{  \left ( G_{i,2}^2 \,  -\,o_i\right ) }{\left [ \alpha_{1}^2 \,  G_{i,1}^2 
 +  \alpha_{2}^2 \,  G_{i,2}^2 
 + \, (\alpha\, o)_i  \right ]} 
 \label{eq:alpha2}
\end{eqnarray}




\subsection{$\mu_{k}$} 
For k=1:
\begin{eqnarray}
0 &= &\sum_i  \sum_{k} Q_i^{t}(k) \, \frac{\partial_{\mu_{k'}} P( p_i, k|\{ \theta_k^{t-1} \})}{P( p_i, k|\{ \theta_k^{t-1} \})}  
\nonumber \\ 
 &= &\sum_{i} Q_{i,1}^t \, \frac{\alpha_{1}^2 \, {\bf C_{1}}^{-1} (p_{i} - \mu_{1})  \, G_{i,1}^2 + \frac{1}{4}\, (1-\alpha_1^2 - \alpha_2^2)\,\frac{G_{i,1}\, G_{i,2} }{ \sum_i G_{i,1} \, G_{i,2}} \, \left ( {\bf C_{1}}^{-1} (p_{i} - \mu_{1}) - \frac{ \sum_i {\bf C_{1}}^{-1} (p_{i} - \mu_{1})\, G_{i,1}\, G_{i,2} }{ \sum_i G_{i,1} \, G_{i,2}}\right )} {\left [ \alpha_{1}^2 \,  G_{i,1}^2 
 + \, \frac{1}{2}(1-\alpha_1^2 - \alpha_2^2)\, \frac{G_{i,1}\, G_{i,2} \, }{ \sum_i G_{i,1} \, G_{i,2}} \right ]} 
\nonumber \\
&& \quad + \, Q_{i,2}^t \, \frac{ \frac{1}{4}\, (1-\alpha_1^2 - \alpha_2^2)\,\frac{G_{i,1}\, G_{i,2} }{ \sum_i G_{i,1} \, G_{i,2}} \, \left ( {\bf C_{1}}^{-1} (p_{i} - \mu_{1}) - \frac{ \sum_i {\bf C_{1}}^{-1} (p_{i} - \mu_{1})\, G_{i,1}\, G_{i,2} }{ \sum_i G_{i,1} \, G_{i,2}}\right )} {\left [ \alpha_{2}^2 \,  G_{i,2}^2 
 + \, \frac{1}{2}(1-\alpha_1^2 - \alpha_2^2)\, \frac{G_{i,1}\, G_{i,2} \, }{ \sum_i G_{i,1} \, G_{i,2}} \right ]}  
 \nonumber \\
  &= &\sum_{i} Q_{i,1}^t \, \frac{\alpha_{1}^2 \, {\bf C_{1}}^{-1} (p_{i} - \mu_{1})  \, G_{i,1}^2 + \frac{1}{4}\, (\alpha\, o)_i  \, {\bf C_{1}}^{-1} \left (  (p_{i} - \mu_{1}) -  \sum_i  (p_{i} - \mu_{1})\, o_i\right )} {\left [ \alpha_{1}^2 \,  G_{i,1}^2 
 + \, \frac{1}{2}(\alpha\, o)_i \right ]} 
\nonumber \\
&& \quad + \, Q_{i,2}^t \, \frac{ \frac{1}{4}\, (\alpha\, o)_i  \, {\bf C_{1}}^{-1} \left (  (p_{i} - \mu_{1}) -  \sum_i  (p_{i} - \mu_{1})\, o_i\right )} {\left [ \alpha_{2}^2 \,  G_{i,2}^2 
 + \, \frac{1}{2}(\alpha\, o)_i  \right ]}  
 \nonumber \\
 && {\rm Replacing \, }
 Q_{i,1}^t \, {\rm and} \, 
  Q_{i,2}^t \, {\rm from\, equations\, } \eqref{eq:E-quantum-step-two-nocos}
\nonumber \\
&= &\sum_{i}  \, \frac{\alpha_{1}^2 \, {\bf C_{1}}^{-1} (p_{i} - \mu_{1})  \, G_{i,1}^2 + \frac{1}{2}\, (\alpha\, o)_i  \, {\bf C_{1}}^{-1} \left (  (p_{i} - \mu_{1}) -  \sum_i  (p_{i} - \mu_{1})\, o_i\right )} {\left [ \alpha_{1}^2 \,  G_{i,1}^2 
 +  \alpha_{2}^2 \,  G_{i,2}^2 
 + \, (\alpha\, o)_i  \right ]} 
\nonumber \\
&= &\sum_{i}  \, {\bf C_{1}}^{-1} (p_{i} - \mu_{1}) \, \frac{\alpha_{1}^2 \,   \, G_{i,1}^2 + \frac{1}{2}\, (\alpha\, o)_i  } {\left [ \alpha_{1}^2 \,  G_{i,1}^2 
 +  \alpha_{2}^2 \,  G_{i,2}^2 
 + \, (\alpha\, o)_i \right ]} 
\nonumber \\
& & -\, \left ( \sum_i  {\bf C_{1}}^{-1} (p_{i} - \mu_{1})\, o_i \right ) \, \sum_{i}  \,   \frac{ \frac{1}{2}\, (\alpha\, o)_i  \,} {\left [ \alpha_{1}^2 \,  G_{i,1}^2 
 +  \alpha_{2}^2 \,  G_{i,2}^2 
 + \, (\alpha\, o)_i \right ]} 
\nonumber \\
&= &\sum_{i}  \, {\bf C_{1}}^{-1} (p_{i} - \mu_{1}) \, \left [  \frac{\alpha_{1}^2 \,   \, G_{i,1}^2 + \frac{1}{2}\, (\alpha\, o)_i  } {\left [ \alpha_{1}^2 \,  G_{i,1}^2 
 +  \alpha_{2}^2 \,  G_{i,2}^2 
 + \, (\alpha\, o)_i  \right ]} \, - \, o_i
 \sum_{j=1}^N  \,   \frac{ \frac{1}{2}\, (\alpha\, o)_j \,} {\left [ \alpha_{1}^2 \,  G_{j,1}^2 
 +  \alpha_{2}^2 \,  G_{j,2}^2 
 + \, (\alpha\, o)_j \right ]} \right ]
\end{eqnarray}
Thus, 
 \begin{eqnarray}
 \mu_1 = \frac{\sum_{i} F_{i,1} \, p_{i}}{\sum_{i} F_{i,1}}\qquad {\rm and} \qquad 
 \mu_2 =\frac{\sum_{i} F_{i,2} \, p_{i}}{\sum_{i} F_{i,2}}
 \label{eq:mu-final}
 \end{eqnarray}
where 
 \begin{eqnarray}
F_{i,k} &=&    \frac{\alpha_{k}^2 \,   \, G_{i,k}^2 + \frac{1}{2}\, (\alpha\, o)_i  } {\left [ \alpha_{1}^2 \,  G_{i,1}^2 
 +  \alpha_{2}^2 \,  G_{i,2}^2 
 + \, (\alpha\, o)_i  \right ]} \, - \, o_i
 \sum_{j=1}^N  \,   \frac{ \frac{1}{2}\, (\alpha\, o)_j \,} {\left [ \alpha_{1}^2 \,  G_{j,1}^2 
 +  \alpha_{2}^2 \,  G_{j,2}^2 
 + \, (\alpha\, o)_j \right ]}  \nonumber \\
 &=&  Q_i(k)\, - \, o_i
 \sum_{j=1}^N  \,   \frac{ \frac{1}{2}\, (\alpha\, o)_j \,} {\left [ \alpha_{1}^2 \,  G_{j,1}^2 
 +  \alpha_{2}^2 \,  G_{j,2}^2 
 + \, (\alpha\, o)_j \right ]} 
\label{eq:mu-def}
\end{eqnarray} 

\subsection{${\bf C_{k}}$}
 We  focus on $\Lambda_k={\bf C_{k}}^{-1}$. 
Recall the maximum likelihood  gives the equation
 \begin{eqnarray}
 0 &=& \sum_i  \sum_k Q_i^{t}(k) \ \frac{\partial_{\Lambda_{k'}} P( p_i, k|\{ \theta_k^{t-1} \})}{P( p_i, k|\{ \theta_k^{t-1} \})}
 \label{eq:maxlikelihood}
 \end{eqnarray}
 Let us investigate $\partial_{\Lambda_{k'}} P( p_i, k|\{ \alpha_k^{t-1},\theta_k^{t-1} \})$ and using equation \eqref{eq:E-quantum-step-two-nocos} 
 we conclude: \\
 \begin{eqnarray}
\partial_{\Lambda_{k'}} P(p_i,k|\{\alpha_k, \theta_k\}) &=& \delta_{k',k} \, \left (
2\, \alpha_{k}^2 \, G_{i,k}  \, \partial_{\Lambda_{k}} G_{i,k} 
 + \frac{1}{4} \, ( 1 -  \alpha_{k}^2 - \, \alpha_{\not k}^2 ) \,
 \frac{G_{i,k} \, G_{i, \not k}} {\sum_i G_{i,k} G_{i, \not k}}\, [\frac{\partial_{\Lambda_{k}} \, G_{i,k}}{G_{i,k}} \, - \frac{ \sum_{i} \,\partial_{\Lambda_{k}} G_{i,k} \, G_{i,\not k}} { \sum_i G_{i,k} G_{i, \not k}}]
 \right ) \nonumber \\
 &+&  \delta_{k',k} \,\left (\frac{1}{4} \, ( 1 -  \alpha_{k}^2 - \, \alpha_{\not k}^2 )
 \frac{G_{i,k} \, G_{i, \not k}} {\sum_i G_{i,k} G_{i, \not k}}\, [\frac{\partial_{\Lambda_{k}} \, G_{i,k}}{G_{i,k}} \, - \frac{ \sum_{i} \partial_{\Lambda_{k}} G_{i,k} \, G_{i,\not k}} { \sum_i G_{i,k} G_{i, \not k}}] \right )\nonumber \\
 &=& \delta_{k',k} \, \left (
2\, \alpha_{k}^2 \, G_{i,k}  \, \partial_{\Lambda_{k}} G_{i,k} 
 + \frac{1}{4} \, (\alpha\, o)_i \, \left [\frac{\partial_{\Lambda_{k}} \, G_{i,k}}{G_{i,k}} \, - \frac{ \sum_{i} \,\partial_{\Lambda_{k}} G_{i,k} \, G_{i,\not k}} { \sum_i G_{i,k} G_{i, \not k}}\right ]
 \right ) \nonumber \\
 &+&  \delta_{k',k} \,\left (\frac{1}{4} \, (\alpha\, o)_i  \, \left [\frac{\partial_{\Lambda_{k}} \, G_{i,k}}{G_{i,k}} \, - \frac{ \sum_{i} \partial_{\Lambda_{k}} G_{i,k} \, G_{i,\not k}} { \sum_i G_{i,k} G_{i, \not k}}\right ] \right )
\label{eq:partial-cpq-P}
\end{eqnarray}
Developing it for the different $(k,k')$ cases, and replacing   $\partial_{\Lambda_{k}}  G_{i,k}= G_{i,k} \,  \partial_{\Lambda_{k}}  \log G_{i,k}   $, we get for k = 1:
\begin{eqnarray}
\quad 0  &=&\sum_i \, Q_{i,1}^t \, \frac {2 \, \alpha_{1}^2 \, G_{i,1}^2 \, (\partial_{\Lambda_{1}}  \, \log G_{i,1}) + \frac{1}{4} \, (\alpha\, o)_i \, [(\partial_{\Lambda_{1}} \log G_{i,1}) - \frac {\sum_{i} G_{i,1} ( \partial_{\Lambda_{1}} \log G_{i,1})\,  G_{i,2}}{ \sum_i G_{i,1} \, G_{i,2}}]} {\alpha_{1}^2 \, G_{i,1}^2 + \frac{1}{2} (\alpha\, o)_i }\nonumber \\
&& + \, 
\sum_{i} Q_{i,2}^t \frac{\frac{1}{4} \, (\alpha\, o)_i  \,[(\partial_{\Lambda_{1}} \log G_{i,1}) - \frac{\sum_{i} G_{i,1} (\partial_{\Lambda_{1}} \log G_{i,1})\,  G_{i,2}}{ \sum_i G_{i,1} \, G_{i,2}}]}{\alpha_{2}^2 G_{i,2}^2 + \frac{1}{2} (\alpha\, o)_i }\nonumber \\
&=&
\sum_i \, \frac {2 \,  \alpha_{1}^2 \, G_{i,1}^2 \, (\partial_{\Lambda_{1}}  \log G_{i,1}) + \frac{1}{2} \, (\alpha\, o)_i  \,[(\partial_{\Lambda_{1}} \log G_{i,1}) - \frac {\sum_{i} G_{i,1} (\partial_{\Lambda_{1}} \log G_{i,1})\,  G_{i,2}}{ \sum_i G_{i,1} \, G_{i,2}}]} {\alpha_{1}^2 \, G_{i,1}^2 + \alpha_{2}^2 \, G_{i,1}^2 + \frac{1}{2}(\alpha\, o)_i }\nonumber \\
&=& 
\sum_i \, \frac {(\partial_{\Lambda_{1}} \, \log G_{i,1})\, (\, 2 \alpha_{1}^2 \, G_{i,1}^2 \, + \frac{1}{2} \, (\alpha\, o)_i  )}{\alpha_{1}^2 \, G_{i,1}^2 + \alpha_{2}^2 \, G_{i,1}^2 + \frac{1}{2}(\alpha\, o)_i } \nonumber \\
&& - \frac{1}{2} \,  \frac { \sum_i \, G_{i,1} \, (\partial_{\Lambda_{1}} \log G_{i,1}) \, G_{i,2}}{\sum_i G_{i,1} G_{i,2}} \, \sum_{i} \frac {\, (\alpha\, o)_i }{\alpha_{1}^2 \, G_{i,1}^2 + \alpha_{2}^2 \, G_{i,1}^2 + \frac{1}{2}(\alpha\, o)_i }\nonumber \\
\end{eqnarray}

Equation for k = 1 becomes the following
\begin{eqnarray}
0 &=& \sum_i  (\partial_{\Lambda_{1}} \log G_{i,1})\, \left [ \frac{(\, 2 \alpha_{1}^2 \, G_{i,1}^2 \, + \frac{1}{2} \, (\alpha\, o)_i  )}{\alpha_{1}^2 \, G_{i,1}^2 + \alpha_{2}^2 \, G_{i,1}^2 + \frac{1}{2}(\alpha\, o)_i} - \, o_{i} \, \sum_{j} \frac {\, \frac {1}{2} \,(\alpha\, o)_j }{\alpha_{1}^2 \, G_{j,1}^2 + \alpha_{2}^2 \, G_{j,1}^2 + \frac{1}{2}(\alpha\, o)_j } \right] \nonumber \\
&=& \sum_i  (\partial_{\Lambda_{1}} \log G_{i,1})\, R_{i,1}
\end{eqnarray}
 and analogously for $k=2$. 
 Recall that  $G_{i,k} = \frac{1}{\sqrt{Z_k}}\,  e^{-  \frac{1}{4}\, (p_i -\mu_k)\, \Lambda_k  \, (p_i-\mu_k) }$ where $\sqrt{Z_k} = (2\pi)^{\frac{d}{4}}|\Lambda_k|^{-\, \frac{1}{4}}$ and  so 

\begin{eqnarray}
\partial_{\Lambda_{k}}\log G_{i,k} &=& 
\frac{1}{4}
\partial_{\Lambda_{k}}\log |\Lambda_{k}|\, - \frac{1}{4}\, (p_i -\mu_k)^T\, (p_i-\mu_k)
\nonumber \\
 &=&\frac{1}{4}\, \frac{1}{|\Lambda_{k}|}\, \partial_{\Lambda_{k}}|\Lambda_{k}|
 \, - \frac{1}{4}\, (p_i -\mu_k)\, (p_i-\mu_k)^T
\nonumber \\
 &=&\frac{1}{4}\, \frac{1}{|\Lambda_{k}|}\, |\Lambda_{k}|  \, \Lambda_{k}^{-1} 
 \, - \frac{1}{4}\, (p_i -\mu_k)\, (p_i-\mu_k)^T\nonumber \\
&=& \frac{1}{4}\, \Lambda_{k}^{-1} 
\, - \frac{1}{4}\, (p_i -\mu_k)\, (p_i-\mu_k)^T
\nonumber \\
&=& \frac{1}{4}\, \left [{\bf C}_{k}
\, - \, (p_i -\mu_k)\, (p_i-\mu_k)^T
\right ]
\end{eqnarray}

Thus,
\begin{eqnarray}
0 &=&\sum_i \,  \left [{\bf C}_{1} \, - (p_i -\mu_1)\, (p_i-\mu_1)^T   \right ]\, \, R_{i,1}^t  \nonumber \\
0 &=&\sum_i \,  \left [{\bf C}_{2} \, - (p_i -\mu_2)\, (p_i-\mu_2)^T   \right ]\, \, R_{i,2}^t  \nonumber \\
& {\rm or} & \nonumber \\
{\bf C}_{k} &=& \frac{\sum_i R_{i,k}^t \,  (p_i -\mu_k)\, (p_i-\mu_k)^T}{\sum_i R_{i,k}^t}
\label{eq:covariance}
\end{eqnarray}
where 
\begin{eqnarray}
R_{i,k}^t &=&  \left [ \frac{(\, 2 \alpha_{k}^2 \, G_{i,k}^2 \, + \frac{1}{2} \, (\alpha\, o)_i  )}{\alpha_{1}^2 \, G_{i,1}^2 + \alpha_{2}^2 \, G_{i,1}^2 + \frac{1}{2}(\alpha\, o)_i} - \, o_{i} \, \sum_{j} \frac {\, \frac {1}{2} \,(\alpha\, o)_j }{\alpha_{1}^2 \, G_{j,1}^2 + \alpha_{2}^2 \, G_{j,1}^2 + \frac{1}{2}(\alpha\, o)_j } \right] \nonumber \\
&=& F_{i,k}^t+\frac{ \alpha_{k}^2 \, G_{i,k}^2 }{\alpha_{1}^2 \, G_{i,1}^2 + \alpha_{2}^2 \, G_{i,1}^2 + \frac{1}{2}(\alpha\, o)_i}
 \label{eq:covar-def}
\end{eqnarray}
\\
For the classical case, $\phi = \frac{\pi}{2}$, \, $R_{i,k}^t=2\, Q_{i,k} $\\
 $${\bf C_k} =\frac{ \sum_i Q_{i,k}\, (p_i-\mu_k)(p_i-\mu_k)^{T}}{ \sum_i Q_{i,k}}$$

\section{Brief Derivation of the EM Method}
\label{sec:Derivation-EM}
Here is a brief derivation of the EM method.  We denote
the parameters $\alpha_k, \theta_k=\{  {\bf C}_k, \mu_k \}$ and the set of data by $\{ p_i\}$. Thus,

\begin{eqnarray} 
P(  \{ p_i \} | \{\theta_k\})= \prod_i P(  p_i |\{\theta_k\})= \prod_i \left ( \frac{P( p_i  , k|\{ \theta_k \})}{P(k| p_i  ,\{ \theta_k \})} \right )
\label{eq:E-M-starting-proof}
\end{eqnarray}

 Given an estimate value $\{ \theta_k \}$, the likelihood is 
\begin{eqnarray}
L( \{ p_i \} |\{ \theta_k\}) &=&  
\log P(  \{ p_i \} | \{\theta_k\}) 
\nonumber \\
&=& \sum_i\, \log P( p_i|\{ \theta_k \})
\nonumber \\
&=& \sum_i\, \log \sum_k P( p_i, k|\{ \theta_k \})
\nonumber \\
&& {\rm introducing\, } (\, Q_i(k)  \ge 0 \, {\rm and} \, \sum_k Q_i(k) =1\, )
\nonumber \\
&=& \sum_i \, \log \sum_k Q_i(k) \frac{ P( p_i, k|\{ \theta_k \})}{Q_i(k)}
\nonumber \\ 
&& {\rm using \, Jensen's \, Inequality}
\nonumber \\
&\ge& \sum_i  \sum_k Q_i(k) \, \log \frac{ P( p_i, k|\{ \theta_k \})}{Q_i(k)}
\label{eq:E-M-ending-proof}
\end{eqnarray}

Now, for any set of distributions $Q_i(k)$ 
, the formula \eqref{eq:E-M-ending-proof} gives a lower-bound on $L( \{ p_i \} |\{ \theta_k\})$.  If we have some current guess $\{ \theta_k^{t}\}$ of the parameters,  we  make the lower-bound tight at that value of $\theta$.  Thus, we need for the step
involving Jensen's inequality in our derivation above to hold with equality. It happens at $Q_i(k)=P(k| p_i  ,\{ \theta_k \})$. 

Then, we iterate on estimation of $[ P(k|\{ \theta_k \}),\{ \theta_k \} ]$ and $P(k| p_i  ,\{ \theta_k \})$. We index these iterations by $t$. First, at $t=0$ of the iteration procedure, a random set of values $[P^{t=0}(k|\{ \theta_k \}),\theta_k^{t=0}]$ is produced. 
{\small
\bibliographystyle{ieee}
\bibliography{quantum.project}

\begin{thebibliography}{10}\itemsep=-1pt

\bibitem{Malik1998}
S.~Belongie, C.~Carson, H.~Greenspan, and J.~Malik.
\newblock Color- and texture-based image segmentation using em and its
  application to content-based image retrieval.
\newblock In {\em Computer Vision, 1998. Sixth International Conference on},
  pages 675--682, Jan 1998.

\bibitem{Bilmes1998}
J.~Bilmes.
\newblock A gentle tutorial of the em algorithm and its application to
  parameter estimation for gaussian mixture and hidden markov models.
\newblock {\em International Computer Science Institute}, pages 1--281, 1998.

\bibitem{Boykov2001}
Y.~Boykov, O.~Veksler, and R.~Zabih.
\newblock Fast approximate energy minimization via graph cuts.
\newblock {\em IEEE Transactions on Pattern Analysis and Machine Intelligence},
  23(11):1222--1239, Nov 2001.

\bibitem{CELENK1990}
M.~Celenk.
\newblock A color clustering technique for image segmentation.
\newblock {\em Computer Vision, Graphics, and Image Processing}, 52(2):145 --
  170, 1990.

\bibitem{Chen2008}
T.-W. Chen, Y.-L. Chen, and S.-Y. Chien.
\newblock Fast image segmentation based on k-means clustering with histograms
  in hsv color space.
\newblock In {\em Multimedia Signal Processing, 2008 IEEE 10th Workshop on},
  pages 322--325, Oct 2008.

\bibitem{ICIP2015}
M.~Cicconet, D.~Geiger, and M.~Werman.
\newblock Complex-valued hough transforms for circles.
\newblock 2015.
\newblock Quebec City, Canada.

\bibitem{Feynman1971}
R.~Feynman.
\newblock {\em The Feynman Lectures on Physics}, volume~3.
\newblock Addison Wesley, 1971.

\bibitem{Ishikawa1998}
H.~Ishikawa and D.~Geiger.
\newblock Segmentation by grouping junctions.
\newblock In {\em Computer Vision and Pattern Recognition, 1998. Proceedings.
  1998 IEEE Computer Society Conference on}, pages 125--131, Jun 1998.

\bibitem{kim2007}
S.~C. Kim and T.~J. Kang.
\newblock Texture classification and segmentation using wavelet packet frame
  and gaussian mixture model.
\newblock {\em Pattern Recognition}, 40(4):1207 -- 1221, 2007.

\bibitem{Likas2003}
A.~Likas, N.~Vlassis, and J.~J. Verbeek.
\newblock The global k-means clustering algorithm.
\newblock {\em Pattern Recognition}, 36(2):451 -- 461, 2003.
\newblock Biometrics.

\bibitem{K1976}
K.~McLaren.
\newblock The development of the cie (l∗, a∗, b∗)-uniform color space,
  1976.

\bibitem{Srivastava2005}
A.~Srivastava, S.~H. Joshi, W.~Mio, and X.~Liu.
\newblock Statistical shape analysis: clustering, learning, and testing.
\newblock {\em IEEE Transactions on Pattern Analysis and Machine Intelligence},
  27(4):590--602, April 2005.

\bibitem{Tomasi98}
C.~Tomasi and R.~Manduchi.
\newblock Bilateral filtering for gray and color images.
\newblock In {\em Computer Vision, 1998. Sixth International Conference on},
  pages 839--846, Jan 1998.

\end{thebibliography}
}

\end{document}